\documentclass[10pt]{article}

\usepackage{amsmath, amssymb, amsthm, xcolor, fullpage, supertabular}
\usepackage[hyphens]{url}
\usepackage[colorlinks, citecolor=, linkcolor=, urlcolor=]{hyperref}
\usepackage[neverdecrease]{paralist}
\usepackage{mleftright} \mleftright

\usepackage{natbib}
\setcitestyle{round}   % disable for [] citations
\usepackage{graphicx}
\usepackage[all]{xy}\UseComputerModernTips

\usepackage{placeins}   % for floatbarrier
\usepackage{algorithm, algpseudocode}
\usepackage{rotating}    % for landscape figures

\usepackage{microtype}
\DisableLigatures[f]{encoding = *, family = * }

\usepackage{epstopdf}

\graphicspath{ {./figures/}}

\renewcommand*{\equationautorefname}{Equation}
\def\equationautorefname~#1\null{(#1)\null}		% place parentheses around equation number

\newcommand*{\quark}{\setbox0\hbox{$x$}\hbox to\wd0{\hss$\cdot$\hss}}

\newcommand{\todo}[1]{\bgroup\bfseries\color{red}#1\egroup}

\newcommand*{\arXiv}[1]{URL \bgroup\color{blue}\href{https://arxiv.org/abs/#1}{https://arxiv.org/abs/#1}\egroup}
\newcommand*{\doi}[1]{\bgroup\color{blue}\href{https://dx.doi.org/#1}{doi:#1}\egroup}
\newcommand*{\email}[1]{\bgroup\color{blue}\href{mailto:#1}{#1}\egroup}
\renewcommand*{\url}[1]{\bgroup\color{blue}\href{#1}{#1}\egroup}
\newcommand*{\ppara}[1]{\noindent\textbf{\textsf{#1}}\,\,}
\newcommand{\eq}{\! = \!}
\newcommand{\iid}{\stackrel{_{iid}}{\sim}}
\newcommand{\argmax}{\operatornamewithlimits{argmax}}
\newcommand{\arXivOmit}[1]{}

\begin{document}

\title{\bf Model-based clustering in very high dimensions via adaptive projections}

\author{%
	B.\ Taschler\footnote{Statistics and Machine Learning, German Center for Neurodegenerative Diseases (DZNE), 53127 Bonn, Germany, \email{bernd.taschler@dzne.de}}
	\and
	F.\ Dondelinger\footnote{Centre for Health Informatics, Computing and Statistics, University of Lancaster, Lancaster, LA1 4YG, United Kingdom, \email{f.dondelinger@lancaster.ac.uk}}
	\and
	S.\ Mukherjee\footnote{Statistics and Machine Learning, German Center for Neurodegenerative Diseases (DZNE), 53127 Bonn, Germany, \email{sach.mukherjee@dzne.de}}
}

\date{\today}

\maketitle

\begin{abstract}
	\ppara{Abstract:}
	Mixture models are a standard approach to dealing with heterogeneous data with non-i.i.d. structure. However, when the dimension $p$ is large relative to sample size $n$ and where either or both of means and covariances/graphical models may differ between the latent groups, mixture models face statistical and computational difficulties and currently available methods cannot realistically go beyond $p \! \sim \! 10^4$ or so.
We propose an approach called {\it Model-based Clustering via  Adaptive Projections} ({\it MCAP}). Instead of estimating mixtures in the original space, we work with a low-dimensional representation obtained by linear projection. 
The projection dimension itself plays an important role and governs a type of bias-variance tradeoff with respect to recovery of the relevant signals. 
MCAP sets the projection dimension automatically in a data-adaptive manner, using a proxy for the assignment risk. 
Combining a full covariance formulation with the adaptive projection allows  detection of  both mean and covariance signals in very high dimensional problems.
We show real-data examples in which covariance signals are reliably detected in problems with $p \! \sim \! 10^4$ or more, and simulations going up to $p = 10^6$.
In some examples, MCAP performs well even when the mean signal is entirely removed, leaving differential covariance structure in the high-dimensional space as the only signal.
Across a number of  regimes, MCAP performs as well or better than a range of existing methods, including a recently-proposed $\ell_1$-penalized approach; and performance remains broadly stable with increasing dimension. 
MCAP can be run ``out of the box" and is fast enough for interactive use on large-$p$ problems using standard desktop computing resources.  
	
	\smallskip
	
	\ppara{Keywords:} 
	Mixture models, linear projections, clustering, graphical models, high-dimensional data
	
\end{abstract}

% %%%%%%%%%%%%%%%%%%%%%%
\section{Introduction}
\label{sec:intro}

% mixture models
Consider an $n \! \times \! p$ data matrix $X \eq [x_1 \ldots x_n]^{\mathrm{T}}$, where $p$ is the dimension and $n$ the sample size. Suppose the data are not i.i.d.\ but rather conditionally independent and identically distributed (i.i.d.), with 
each data point $x_i$ having a corresponding discrete latent variable $z_i \! \in \! \{ 1 \ldots K \}$ indicating group membership. A standard approach in this setting is to model the data in each group $k$ via a (group-specific) $p$-dimensional distribution  $g_k$. This gives the marginal over $x$ as a mixture  of the form $p(x) \eq \sum_k \pi_k g_k(x)$, where $\pi_k \eq P(z \eq k)$ are (marginal) group membership probabilities. 
%Estimation in this setting is very well studied in the low- to moderate-dimensional case, but the large $p$ case remains challenging. 

In this paper we propose an approach called {\it Model-based Clustering via  Adaptive Projections}  (MCAP) that 
models data of this kind, 
using linear projections to cope with high dimensionality. 
Real-world high-dimensional data are often characterized by a nontrivial covariance structure that is not necessarily identical across groups and that cannot be assumed to be diagonal. We are therefore   interested in detecting signals not only in cluster-specific means, but also in cluster-specific covariances. 
Suppose $\mu_k, \Sigma_k$ are respectively the (population) mean and covariance matrix for group $k$. We say that a  problem has a {\it mean signal} if the $\mu_k$'s are non-identical across all groups, and, analogously, a {\it covariance signal} if the $\Sigma_k$'s are non-identical. A specific problem may have either or both types of signal. 

Nontrivial differences in covariance composition are a feature of many group-structured real data. Recent work has looked at high-dimensional estimation \citep{Danaher2014TheClasses} and testing for differential covariance structure \citep{Stadler2017Two-sampleDimensions} when group indicators are known. 
For example in biology, gene-gene covariances or graphical models may differ between disease subtypes due to underlying biological differences; or in retail data, correlation patterns between attributes might differ by customer strata. Such signals, if present, may be  useful for identifying clusters and in addition may in themselves be of scientific or practical interest. 
Furthermore, due to Simpson's paradox \citep[see e.g.][]{Pearl2016CausalPrimer}, direct learning of covariances or graphical models  without accounting for the latent structure can give completely incorrect results, even asymptotically.

%Furthermore, it is possible that covariance signals could in some cases be as strong or stronger than mean signals.
% (naturally, qualifying this would require a suitable general definition of signal strength, e.g. from an information or decision-theoretic point of view).

Many standard clustering methods, such as K-means, are aimed at detecting mean signals. In contrast, suitably formulated model-based clustering methods can in principle detect both mean and covariance signals (at least in the large sample setting) since they fully model the underlying distributions. 
Mixture models and model-based clustering have been extensively studied \citep[see e.g.][]{McLachlan2000FiniteModels} and many excellent tools are available, such as the widely-used \texttt{mclust} software \citep{Fraley2012MclustEstimation}. However, the higher dimensional domain remains challenging, particularly if covariance signals are of interest and  when $p \gg n$.
For full covariance Gaussian mixtures, the key problem is the high-dimensionality of the parameter space. Learning large covariance matrices is already statistically and computationally nontrivial in the non-latent case \citep{Friedman2008SparseLasso,Zhao2012TheR}, and these difficulties are exacerbated in the latent variable setting \citep{Zhou2009PenalizedMatrices,Stadler2013PenalizedModels,Stadler2017MolecularStudy}. 
%In the interests of tractability, strong assumptions are often imposed on covariance structures, such as diagonal covariances or identical covariances across groups, but these can weaken or eliminate the ability to detect covariance signals. 

% previous work I: penalized methods
Penalized approaches have been proposed for  mixtures in high dimensions \citep{Zhou2009PenalizedMatrices,Stadler2017MolecularStudy}. These methods extend sparse penalized estimation---as familiar from the lasso, graphical lasso and related methods---to mixture modelling.
This can be  effective  and current methods can deal with relatively large $p$ problems, whilst coping with covariance matrices with perhaps millions of entries. 
However, despite these advances, the burden of full scale high-dimensional modeling in this setting is substantial. For example, in expectation-maximization (EM)-type algorithms, the relevant high-dimensional estimators must be invoked for each cluster and at each EM iteration. At present, due to these factors, this class of methods remains restricted in terms of $p$  and their empirical performance on large problems, including the setting of tuning parameters etc., remains incompletely studied. 
Below, we present extensive empirical results comparing MCAP to the $\ell_1$-penalized mixture model \texttt{mixGlasso} \citep{Stadler2017MolecularStudy}, as well as several traditional clustering methods, including \texttt{mclust} \citep{Scrucca2016MclustModels}. 
MCAP extends to much larger $p$ than currently available penalized approaches. However, even in most examples where it was feasible to run \texttt{mixGlasso} in the original high-dimensional space, MCAP outperforms it whilst requiring a small fraction of the computational effort.

% prev work II: projection, more for mean signals, setting q is difficult, random vs PCA-type projections
Several classical, mean- or distance-based methods (e.g.\ K-means) scale relatively well to large $p$, as do spectral clustering  methods \citep{vonLuxburg2006AClustering}. However, these approaches are aimed at detecting mean signals. 
Another approach uses projections to reduce the dimension of the data, followed by a clustering step on the projected data. 
This has long been a standard heuristic approach in diverse applications and for random projections has been systematically studied, informed by the well-known Johnson-Lindenstrauss lemma \citep{Dasgupta2003AnLindenstrauss,Ailon2009TheNeighbors}, which considers the behaviour of relative distances under random projections. Here, the literature has focused mainly on the case of mean signals, and in practice sensitivity to the projection dimension can be an issue.

% what we do
Our approach belongs to the latter class of projection-based methods but we focus attention on (i) both mean and covariance signals and (ii) on setting the projection dimension automatically.
Specifically, we project  from $p$ to $q \! < \! p$ dimensions and then carry out mixture modelling in $q$ dimensions. 
%This renders mixture modelling feasible, since the mixture need only be fitted on the low dimensional data. 
This allows the use of full covariance models in the second step, since the dimension is then sufficiently low to control the variance of the estimator.

MCAP is motivated from an assignment risk point of view. Assignment risk is related to the notion of generative vs.\ discriminative learning in high-dimensional supervised problems \citep{Prasad2017OnModels}. The key idea is a type of bias-variance tradeoff, controlled by the target dimension $q$.
If $q$ is too small, the relevant signals can be lost. On the other hand, if $q$ is too large, the statistical (and computational) cost increases rapidly, rendering analysis ineffective and/or intractable. 
A main contribution of this work is the setting of $q$ in a data-adaptive manner, using a stability-based score derived from (parallelizable) clustering of subsets of the original data.

For the projection step, MCAP can in principle use any projection with a specified target dimension. We consider principal component analysis (PCA) and random projections, including dense ({\it MCAP-RP-Gauss}) as well as sparse variants ({\it MCAP-RP-Achl}, {\it MCAP-RP-Li}). We find that PCA projections with the target dimension $q$ set adaptively are  effective in a wide range of problems and propose this as a default. The formulation is applicable to very high dimensional problems with $p$ in the tens or hundreds of thousands and can be run out-of-the-box.

A particularly challenging case is that of high-dimensional problems where the covariance signal is important. We show that our proposed approach is effective in settings where  $p$ is much too large for direct mixture modeling (even by state-of-the-art penalized approaches) and can even be successfully applied in cases where the {\it only} signal is in the covariance. %In contrast, random projections are effective in a narrower range of circumstances (see also Discussion).
%The approaches we propose provide, to the best of our knowledge, the only  practical way to perform mean-and-covariance-based clustering in very high dimensions. 
To give an idea of practical efficiency, cluster assignments and responsibilities for a problem with $p \eq 10^5$ can be obtained within a few minutes on a standard laptop without parallelization, including projection and adaptive setting of $q$.

The remainder of the paper is organized as follows. First, we introduce notation and describe the methods. The key elements are projection followed by full-covariance   mixture modeling coupled with adaptive setting of the projection dimension. 
We consider a range of empirical examples, mostly based on real data, but with cluster assignments known on the basis of clear external information. This allows us to consider real-world covariance structures and at the same time rigorously evaluate the methods against a gold standard. 
We also show results on high-dimensional graphical model recovery and when data follow a simpler model with no differential covariance structure.
We close with a discussion of our results and point to some directions for future work.

\section{Methods}
\label{sec:methods}

\subsection{Notation}
Let $X \eq [x_1 \ldots x_n]^{\mathrm{T}}$ denote an $n{\times}p$ data matrix and $z_i \! \in \! \{1, \ldots, K \}$ denote latent variables corresponding to the data points and indicating the underlying (unknown) group membership. 
Where useful to emphasize the sample size, we write $X_n$ for the data matrix.
We generically denote a dimension reducing function as $T_q : \mathbb{R}^p \rightarrow \mathbb{R}^q$ and refer to the  dimension $q$ to which the data is reduced as the {\it target dimension}.
%A $p \times q$ PCA projection matrix is denoted $W_{\mathrm{PCA}}$, i.e. using PCA the reduced dimension $n \times q$ data matrix is $X_{\mathrm{PCA}} \eq X W_{\mathrm{PCA}}$. Similarly, for a random projection the $p \times q$ projection matrix is denoted $W_{\mathrm{rand}}$ and the projected data  $X_{\mathrm{rand}} \eq X W_{\mathrm{rand}}$.
Where useful we indicate the target dimension in superscript; that is, $X^{(q)}=T_q(X)$ generically denotes a $q$-dimensional reduction of the data. A multivariate normal distribution with mean $\mu$ and covariance $\Sigma$ is denoted $\mathrm{N}(\mu,\Sigma)$. Parameters of a Gaussian mixture model with $K$ components are denoted 
$\theta = ( \pi_k, \mu_k,\Sigma_k)_{k=1..K}$, with dimensionality $p$ or $q$, as will be clear from the context.

\subsection{Assignment risk in mixture modelling}
\label{sec:risk}
Consider a Gaussian mixture model with $K$ components, i.e.\ the model with marginal density 
\begin{equation}
p(x) = \sum_{k=1}^K \pi_k \, \mathrm{N}(x \mid \mu_k,\Sigma_k).
\end{equation}
The ($p$-dimensional) model parameters are $\theta = ( \pi_k, \mu_k,\Sigma_k)_{k=1..K}$. 
%Let $\hat{\theta}_{\mathrm{MLE}}$ denote the maximum likelihood estimator (MLE). We note that due to computational considerations arising from the associated optimization problem, in general the MLE cannot be obtained in practice; rather, the output of a standard EM-based estimator is a local optimum. We use $\hat{\theta}$  to denote such an estimator and reserve the subscript MLE for the theoretical true maximum likelihood estimate.
A fitted mixture model allows each data point to be assigned to one of the $K$ groups. These assignments are estimates of the latent variables $Z_n \eq [z_1 \ldots z_n]^{\mathrm{T}}$. 

It will be instructive to consider the risk for the assignment problem rather than the parameter itself. 
Thus, let $C(X_n; \theta) \in \{1, \ldots, K \}^n$ be assignments for the data vectors in $X_n$, obtained using the parameter $\theta$ (for example, assigning each point to the most likely component $k$). The finite sample {\it assignment risk} associated with an estimator $\hat{\theta}$ is
\begin{equation}
R_n(\hat{\theta}) = \mathbb{E}[ L(Z_n, C(X_n; \hat{\theta}(X_n))],
\end{equation}
where $L$ is a loss function and the expectation is with respect to the joint distribution of the latent assignments and observed data. Due to the label switching problem, a suitable loss function should be invariant to label permutation.
The loss function of Binder \citep{Binder1978BayesianAnalysis,Fritsch2009ImprovedMatrix} is one example (this is essentially one minus the Rand index) and related ideas, linking clustering to the latent prediction problem, are discussed in \citet{Tibshirani2005ClusterStrength}.
Furthermore, denote the risk under the true data-generating parameter $\theta^*$ as 
\begin{equation}
R^*_n = \mathbb{E}[ L(Z_n, C(X_n; \theta^*))].
\end{equation}
This is analogous to the Bayes' risk in discriminant analysis. For an estimator $\hat{\theta}$, the excess assignment risk is $R_n(\hat{\theta}) - R^*_n$.

%Under mild conditions the MLE is consistent, that is $\hat{\theta}_{\mathrm{MLE}}(X_n) \rightarrow \theta^*$, where $\theta^*$ is the true value of the parameter. As a consequence $R_n(\hat{\theta}_{\mathrm{MLE}}) \rightarrow R^*_n$ asymptotically, i.e. for large samples the excess assignment risk under the MLE goes to zero.

Next, consider the behaviour of a dimension reduction  $T_q \in \mathcal{T}, T_q: \mathbb{R}^p \rightarrow \mathbb{R}^q$ followed by estimation in the low-dimensional space. The assignment risk is 
\begin{equation}
R_{n,\mathcal{T},q}(\hat{\theta}) = \mathbb{E}[ L(Z_n, C(X_n; \hat{\theta}^{(q)}(T_q(X_n)))],
\end{equation}
where the subscripts indicate that the risk depends not only on $n$ and the estimator $\hat{\theta}$ but also the specific class $\mathcal{T}$ of the projection function and the target dimension $q$.
Note that under dimension reduction, the Gaussian parameters $\hat{\theta}^{(q)}$ refer to the projected problem in $q$ rather than $p$ dimensions. Nevertheless, the assignment risk remains  well-defined and comparable, since it is  the same assignment problem that is being solved, albeit via a transformation of the data.

Now consider  the behavior of the assignment risk for varying $q$. 
In finite samples, large $q$ may lead to poor assignments and high risk due to the fact that  as $q$ grows, the variance $\mathrm{Var}(\hat{\theta}^{(q)}(T_q(X_n)))$ of the estimator increases and, in case of full covariance models and maximum likelihood estimation, does so  rapidly. 
On the other hand, for small $q$, the excess risk may be large even asymptotically since the transformation $T_q$ in general loses information.

From this point of view, the target dimension $q$ can be thought of as a tuning parameter that governs a bias-variance-type tradeoff: A larger $q$ gives lower excess risk asymptotically (low bias), but in finite samples the variance may be very high. In contrast, smaller $q$ gives low variance estimates, but may not guarantee low excess risk, even asymptotically (high bias).

In general, the optimal $q$ will depend on (unknown) details of the data generating model. For example, in settings where $K=2$ and the mean group difference is large, using a small $q$ in projections via PCA will typically suffice, since the first PCs will, with high probability, capture the difference $\Delta \mu = \mu_1 - \mu_2$. On the other hand, if the mean difference is smaller and the covariance signal important, a larger $q$ may be needed. Furthermore, details of the estimator $\hat{\theta}$ will be influential, as well as overall sample size. 

In light of the above arguments, we propose to treat the target dimension as a tuning parameter to be set in a data-adaptive manner, as we discuss below.

\subsection{Adaptive projections}
In order to specify a practical adaptive  scheme, we need to specify the class of functions $\mathcal{T}$, an estimator $\hat{\theta}$ and a way to set  $q$ using the available data. 
For $\mathcal{T}$, we consider linear projections, specifically PCA and random projections. 
For the estimator, we use classical expectation-maximisation (EM) for a Gaussian mixture model with entirely unconstrained covariances, i.e.\ we allow for full covariances that can differ between components $k$. Hereafter $\hat{\theta}$ refers specifically to this estimator.

One could also use a regularized estimator---such as those proposed by \citet{Zhou2009PenalizedMatrices} or \citet{Stadler2017MolecularStudy}---that would reduce the variance at a given $q$.  However, in our scheme the main work of controlling the bias-variance tradeoff is done by $q$ and we prefer to remove the need for any additional tuning parameters. 
The choice of full covariance is motivated by the need to detect covariance signals and is made  feasible by bounding $q$ (see below) so that $\hat{\theta}$ does not need to be applied to the high-dimensional data directly. 
As we show later in the empirical examples, this strategy is effective in high-dimensional problems and computationally extremely efficient.

Ideally we would like to set $q$ in a way as to minimize the assignment risk $R_{n,\mathcal{T},q}(\hat{\theta})$. 
%We propose to set $q$ automatically. For a class $T_q$ of projections, we would ideally like to minimize the excess risk associated with projection dimension $q$, i.e. set $q$ to minimize $R_{n,T,q}(\hat{\theta})$, where $\hat{\theta}$ is the specific estimator used in the second, mixture modeling step. 
However, this quantity is not available in practice, nor can it be replaced with a direct sample analogue due to the latent nature of the assignments. Instead, we propose to use a stability-based measure of cluster quality as a rough guide to assignment risk. 
Stability-based measures of cluster quality have been studied in the literature \citep{Hennig2007Cluster-wiseStability}. In general terms, these measures quantify the sensitivity of cluster assignments to perturbation of the data. We emphasize that for our purposes it is not required that a stability measure be a very good estimate of risk, only that it is sufficiently indicative to allow an effective target dimension to be chosen.

We now consider in turn the class of projections $\mathcal{T}$, the data-adaptive setting of target dimension $q$ and the estimator for a given $q$, and then summarize the proposed algorithm.

\subsubsection{Projection}
%The main idea is to project the data $X$ from its original space with dimensionality $p$ down to $q$ dimensions and then apply  Gaussian mixture modelling to the projected data $X^{(q)}$. The latter step is done using an unconstrained, full covariance formulation so that covariance signals, if retained after projection, can be detected. This entails two steps, projection and mixture modeling, which we consider in turn.

We consider random and PCA projections. Random projections have been extensively studied in the  context of mean-based clustering, motivated by the Johnson-Lindenstrauss lemma \citep[see e.g.][]{Dasgupta2003AnLindenstrauss}.

Let $W$ denote the projection matrix of size $p \times q$ with entries $w_{jl}$.  We consider the following three random projections: (i) Standard Normal entries $w_{jl} \iid \mathrm{N}(0,1)$; (ii) A sparse variant as proposed in \cite{Achlioptas2003Database-friendlyCoins}. The projection is specified by \autoref{eq:projection} below, with the parameter $s=3$; and (iii) A very sparse variant following \cite{Li2006VeryProjections}, also specified by \autoref{eq:projection}, but with $s=\sqrt{p}$. Entries of the projection matrices are
\begin{equation}
   \label{eq:projection}
    w_{jl}=\sqrt{s}
    \begin{cases}
       \ \ 1, & \text{with prob.}\ \frac{1}{2s} \\
       \ \ 0, & \text{with prob.}\ 1-\frac{1}{s} \\
       -1, & \text{with prob.}\ \frac{1}{2s}.
    \end{cases}
  \end{equation}
These variants of MCAP based on different random projection techniques are subsequently referred to as \textit{MCAP-RP-Gauss}, \textit{MCAP-RP-Achl} and \textit{MCAP-RP-Li}, respectively.

For high-dimensional data with $p \gg n$, we carry out PCA using a kernel-type formulation, in which the eigendecomposition is done on the $n {\times} n$ Gram matrix $X X^{\mathrm{T}}$ rather than the  matrix $X^{\mathrm{T}} X$ \citep[see e.g.][]{Barber2012BayesianLearning}.
In brief, this is done as follows: For the standard approach, the leading  eigenvectors $v$ of interest  satisfy $(X^{\mathrm{T}} X) v = \lambda v$, which implies $X X^{\mathrm{T}} (X v) = \lambda (X v)$. From the latter expression we see that the projected data can be obtained by eigendecomposition of the $n {\times} n$ Gram matrix. 
In the $p>n$ case this  reduces  computational complexity to $O(n^3)$ rather than $O(p^3)$.
%We note that there are a number of statistical concerns with high-dimensional PCA that have been studied using the tools of random matrix theory.
%However, we note that in the present setting, we do not require recovery of the correct PCs, nor even of the correct subspace, but only that the projection retains class discriminatory information. 

\subsubsection{Mixture modelling} 
In $q$-dimensional space, the data matrix is $X^{(q)} = [x^{(q)}_1 x^{(q)}_2 \ldots\ x^{(q)}_n]^{\mathrm{T}}$ and the model is described by
\begin{equation}
p(x^{(q)}_1,  \ldots, x^{(q)}_n) = \prod_{i=1}^n \sum_{k=1}^K \pi_k \, \mathrm{N}(x^{(q)}_i \mid \mu_k, \Sigma_k),
\end{equation}
where $\mu_k, \Sigma_k$ are the ($q$-dimensional) cluster-specific mean and covariance, respectively. Thus, the model parameters are $\theta = \{ \pi_k, \mu_k, \Sigma_k \}_{k=1\ldots K}$.
Let 
\begin{equation}
\hat{\theta}(X^{(q)}) = \{ \hat{\pi}_k, \hat{\mu}_k, \hat{\Sigma}_k \}_{k=1\ldots K}
\end{equation}
 denote the EM estimates of the model parameters.
The  responsibilities $\gamma_{ik}$ for point $x^{(q)}_i$ and group $k$ are then 
\begin{equation}
\gamma_{ik}(\hat{\theta}(X^{(q)})) = \frac{\hat{\pi}_k \, \mathrm{N}(x^{(q)}_i \mid \hat{\mu}_k, \hat{\Sigma}_k)}{\sum_{k'=1}^K \hat{\pi}_{k'} \, \mathrm{N}(x^{(q)}_i \mid \hat{\mu}_{k'}, \hat{\Sigma}_{k'})} \, \, .
\end{equation}
%Let $c: \{1, \ldots, n \} \rightarrow \{1, \ldots, K \}$ be a cluster assignment function. 
%Cluster assignments are obtained from the parameters $\hat{\gamma}_{ik}$ as
%\begin{equation}
%c(i; X^{(q)}) = \argmax_k \hat{\gamma}_{ik}(X^{(q)}),
%\end{equation}
%where the notation emphasizes that the estimates $\hat{\gamma}_{ik}$ are obtained from data $X^{(q)}$.
Finally, the cluster assignments $C$ are
\begin{eqnarray}
\label{eq:cluster_assignments}
C(X; \hat{\theta}(X^{(q)})) & = & [c_1 \ldots c_n]^{\mathrm{T}} \\ \nonumber
c_i & = & \argmax_k \gamma_{ik}(\hat{\theta}(X^{(q)})).
\end{eqnarray}

%Although we do not focus on this aspect in the present paper, parameter estimates in the original $p$-dimensional space could be recovered in a  final step by using the estimated responsibilities  $\gamma_{ik}(\hat{\theta}(X^{(q)}))$ or assignments $c_i$. In the latter case, this would involve $K$ separate estimation problems. In high dimensions, estimation of covariance matrices or graphical models remains challenging and this would require the use of suitable  regularized estimators  

\subsubsection{Data-adaptive setting of target dimension \texorpdfstring{$q$}{q}}
%In the case of mean-based clustering, the question of how the choice of $q$ affects  the clustering signal has been studied theoretically [ref]. For covariance signals, we are not aware of similar results, but the trade-off can be outlined as follows. Whether for random or PCA-type projections, the information content in the projected data increases with  $q$; this is true in a sample sense for PCA and in expectation for completely random projections. On the other hand, larger $q$ leads to higher variance in the second step (mixture modeling), due to the  increase in the statistical dimension. This is essentially a bias-variance trade-off. 

%Stability analyses can be computationally demanding, since models must be re-fit under data perturbation. However, in the present setting all relevant computations are carried out only on the low-dimensional data, making the computations in fact 

%Here, the data are iteratively subsampled and each such subsample is clustered. For any pair of subsamples, a measure of cluster similarity (the Rand index) is computed for the data points in common between the subsamples, and this is averaged over all pairs of subsamples to give the overall measure. 
% details of cluster stability

As noted above, the assignment risk cannot be directly estimated hence we require a surrogate with which to set the target dimension. We propose to use a stability-based measure as described below. In a nutshell, this involves quantifying the stability of cluster assignments under subsampling of the ($q$-dimensional) data. 

\medskip
\noindent
{\bf Additional notation.} 
For a subset $A \subset \{1, \ldots, n \}$ of sample indices, $X_A$ refers to the corresponding data subsample, i.e.\ the data matrix formed by selecting the rows corresponding to the samples $i \in A$.
Then, let $A_b \subset \{1, \ldots, n \}, \, b = 1 \ldots B$, be (overlapping) subsets of the sample indices, sampled at random without replacement. Each of the $B$ subsets is of size $m<n$. The corresponding $m {\times} q$ data matrices are $X^{(q)}_{A_1}, \ldots ,  X^{(q)}_{A_B}$. That is, these are sub-matrices of $X^{(q)}$, formed,  for each subset $b$, by selecting the $m$ rows with indices $A_b$. 

\medskip
\noindent
{\bf Cluster assignment and data subsets.} 
Consider again the cluster assignment function $C$ in \autoref{eq:cluster_assignments}, where the assignments for data $X$ are written as $C(X; \hat{\theta}(\ldots))$. Note that the argument of the estimator $\hat{\theta}$ need not be $X$ itself; i.e., the estimation could be done on different data then the data for which assignments are produced. 
Using this notation, 
$$C(X_A; \hat{\theta}(X_{A'}))\in \{1, \ldots, K \}^{|A|}$$
are cluster assignments of the points with indices in $A$, 
%based on estimates using  data $X$.
% \[C(A; X) =\{ c(i; X) \}_{i \in A} \in \{1, \ldots, K \}^{|A|}\] denote the clustering assignments of the points in $A$ based on estimates obtained from data $X$.
%That is, $C(X_A; \hat{\theta}(X))$ 
while all estimation is performed using data $X_{A'}$, where $A'$ is another index subset.

%If $A \subseteq A'$, the assignments for $i \in A$ are directly given by the EM output (via the responsibilities $\gamma_{ik}$), otherwise an additional assignment step is needed. 

\medskip
\noindent
{\bf Assignment stability.} 
With this notion---of estimation and assignment on potentially different data subsets---in hand, we can construct a simple measure of stability by considering whether the same set of points are assigned in a similar way when estimation is carried out using different data. The similarity of assignments has to be quantified in a way that accounts for the label-switching problem and therefore the Rand index is a natural choice. 

This leads to the formulation below, where stability is quantified by the Rand index between assignments of the same set of points using potentially different parameters. 
Specifically, let $H(\cdot, \cdot)$ denote the Rand Index between a pair of cluster assignments. 
We score cluster stability using the quantity 
\begin{equation}
\label{eq:cluster_stability}
S_q =  \binom{B}{2}^{-1} \sum_{b \neq b'} H \left( C(X^{(q)}_{A_b \cap A_{b'}}; \hat{\theta}(X^{(q)}_{A_b})), C(X^{(q)}_{A_b \cap A_{b'}}; \hat{\theta}(X^{(q)}_{A_{b'}})) \right)
\end{equation}

In the above expression, the cluster assignments being compared are always assignments of the same set of points, namely those whose indices lie in the intersection $A_b \cap A_{b'}$. However, the data used to perform the clustering differ, i.e.\ those with indices $A_b$ in one case and $A_{b'}$ in the other.
Therefore, $S_q$ is the Rand index between {\it shared} data points in subsamples of the data, averaged over all available pairs of subsets.
The intuition is that if estimation variance is high in a way that actually affects assignment, the assignments will be unstable due to the effect of the data perturbation on 
$C$ via $\hat{\theta}$.

\medskip
\noindent
{\bf Computational issues.} 
Computation of the above measure  requires multiple calls to the estimator $\hat{\theta}$. However, after projection, all operations take place only at the level of the reduced data $X^{(q)}$ and, furthermore, each call to the clustering $C$ can be done in embarrassingly parallel fashion. This means that in practice the tuning step can be carried out very efficiently. 
Note also that, in practice, computing $S_q$ does not require an additional assignment step since the required assignments are obtained directly from the EM output on each subset (because $A_b \cap A_{b'} \subseteq A_b$ and 
$A_b \cap A_{b'} \subseteq A_{b'}$).

\medskip
\noindent
{\bf Setting projection dimension $q$.} 
Considering a grid $\mathcal{Q}$ of candidate values, we set $q$ as
\begin{equation}
\hat{q} = \argmax_{q \in \mathcal{Q}} S_q \, .
\end{equation}

We use a coarse grid for $\mathcal{Q}$, with $\max(\mathcal{Q}) = \left\lfloor \sqrt{10n/k} \right\rfloor$, the latter motivated by the total number of model parameters (in the projected space) relative to sample size. In all experiments we set the size of each of the $B$ subsets to $m = \lfloor 0.75 \, n \rfloor$.

%We provide timing information below, but to give a rough idea, for either random projections or PCA, a high-dimensional problem can be run in a short time on a standard laptop in serial, net of all steps, including projection and adaptive setting of $q$. 

\medskip
\noindent
{\bf Algorithm summary.} 
\autoref{alg:MCAP} summarizes the proposed method.
We refer to the proposed method in general as \textit{MCAP}. When discussing specific implementations, variants using PCA or random projections are referred to as {\it MCAP-PCA} or {\it MCAP-RP}, respectively.

\hspace{10pt}
\begin{algorithm}
\caption{MCAP algorithm}
\label{alg:MCAP}
\begin{algorithmic}[1]
	\Require $n {\times} p$ data matrix $X$; Number of groups $K$; Class of projections $\mathcal{T} = \{T_q \}$
    \Ensure Cluster assignments $c_i \in \{1,\dots, K\}$ with \mbox{$i=1 \ldots n$}; Mixture model parameters $\hat{\theta}$
    \ForAll {$q \in \mathcal{Q}$}
	\State Obtain projected data $X^{(q)} \gets T_q(X)$
    	\For {$b=1:B$}
        	\State Sample $A_b \subset \{1, \dots, n\}$, $|A_b|=\lfloor 0.75n \rfloor$
            \State Estimate parameters $\hat{\theta}_b \gets \hat{\theta}(X^{(q)}_{A_b})$
        	\State Obtain assignments $c_b \gets C(X^{(q)}_{A_b}; \hat{\theta}_b)$
        \EndFor              
		\State Compute $S_q \gets S_q(c_1, \dots, c_B)$, using \autoref{eq:cluster_stability}       
	\EndFor    
    \State Obtain optimized $\hat{q} \gets \argmax S_q$, $q \in \mathcal{Q}$
    \State Estimate  model parameters $\hat{\theta}_{\textnormal{out}} \gets \hat{\theta}(T_{\hat{q}}(X))$ \\
    \Return $\hat{\theta}_{\textnormal{out}}$; $C_{\textnormal{out}} \gets C(T_{\hat{q}}(X); \hat{\theta}_{\textnormal{out}})$
\end{algorithmic}
\end{algorithm}

\subsection{Estimation of high-dimensional parameters and graphical models}
\label{sec:param_estimation}
The mixture model parameters in the original high dimensional space are group-specific mean vectors $\mu_k \in \mathbb{R}^p$ and $p {\times} p$ covariance matrices $\Sigma_k = \Omega_k^{-1}$.
If available, the inverse covariance matrices $\Omega_k$ can be used to obtain group-specific Gaussian graphical models. However, the parameter estimates $\hat{\theta}(X^{(q)})$ obtained after projection are  $q$-dimensional. 
When the high-dimensional parameters themselves---rather than only the assignments or responsibilities---are of interest, we recover them in a final step, using the responsibilities to either split or weight the data. We consider two specific schemes, hard and soft assignments. 

\medskip

\noindent
{\bf Hard assignment}.
The data $X$ are split into $K$ disjoint groups using the assignments $C(X; \hat{\theta}(X^{(q)}))$. Let $S_k$ denote the estimated index subsets, i.e.\ $S_k = \{ i : c_i = k \}$. Then the corresponding data subsets are $X_{S_1}, \ldots , X_{S_K}$. 
These $p$-dimensional data can be used to estimate high-dimensional parameters directly, for example, via a call to a regularized estimator such as the  graphical lasso. We focus on the graphical lasso in what follows, but note that similar ideas could be used with other high-dimensional estimators. 

Let $\hat{\Omega}(\cdot)$ denote the graphical lasso estimator. 
Under hard assignment, we estimate the  $p$-dimensional parameters as 
\begin{eqnarray}
\hat{\mu}_k & = & \hat{n}_k^{-1} \sum_{i \in S_k} x_i,\\
\hat{\Omega}_k & = & \hat{\Omega}(X_{S_k}),
\end{eqnarray}
where $\hat{n}_k = | S_k |$  are the estimated group-specific sample sizes.

Using the graphical lasso leads to sparse estimates $\hat{\Omega}_k$. The zero patterns in these $p {\times} p$ matrices then give group-specific graphical models. In examples below, for each call to the graphical lasso, we set the tuning parameter by cross-validation (CV).
Note that the approach described here requires exactly $K$ calls to the graphical lasso and is for this reason feasible for much larger $p$ than direct mixture modelling in the high-dimensional space as, for example, in \citet{Zhou2009PenalizedMatrices} and \citet{Stadler2017MolecularStudy}.

\medskip

\noindent
{\bf Weighted (``soft") estimation.} 
In the previous paragraph, the discrete assignments $C$ were obtained by thresholding the responsibilities $\hat{\gamma}_{ik} = \gamma_{ik}(\hat{\theta}(X^{(q)}))$.
Alternatively, the $\hat{\gamma}_{ik}$'s themselves can be used to perform a weighted estimation of the high-dimensional parameters. This can be thought of as similar to carrying out a single M-step in the high-dimensional space, carrying over the responsibilities obtained from the low-dimensional modelling. For the ($p$-dimensional) means, this gives 
\begin{eqnarray}
\hat{\mu}_k & = & \hat{n}_k^{-1} \sum_{i=1}^n  \hat{\gamma}_{ik} \, x_i, 
\end{eqnarray}
with $\hat{n}_k = \sum\limits_{i=1}^n \hat{\gamma}_{ik}$.
Similarly, the entries of the $p$-vectors of group-specific marginal variances $v_k$ are estimated as 
$$\hat{v}_{kj}  = \hat{n}_k^{-1} \sum_{i=1}^n  \hat{\gamma}_{ik} \, (x_{ij}  - \mu_{kj})^2,$$ 
where for any $p$-vector $b_{\cdot}$ the $j^{\textnormal{th}}$ entry is denoted as $b_{\cdot j}$.
For the inverse covariance matrices we perform the following graphical lasso optimization:
\begin{eqnarray}
\hat{\Omega}_k & = & \argmax_{\Omega_k} \hat{n}_k \left( \log \det \Omega_k - \mathrm{Tr}(U_k \Omega_k)  - \lambda_k \, \| \Omega_k \|_1 \right),
\label{eq:glasso}
\end{eqnarray}
where %$U_k = \frac{1}{n_k} \sum_{i=1}^n \hat{\gamma}_{ik}\,  x_i x_i^{\mathrm{T}}  - \hat{\mu}_k \hat{\mu}_k^{\mathrm{T}}$
$$U_k = \hat{n}_k^{-1} D_k \bigl( \sum_{i=1}^n  \hat{\gamma}_{ik} \,  (x_i -\hat{\mu}_k) (x_i - \hat{\mu}_k)^{\mathrm{T}} \bigr) D_k^{\mathrm{T}}$$ and $$D_k = \mathrm{diag}(\hat{v}_k)^{-1/2}.$$

The matrix $D_k$ sets  the variables to unit scale. This is in effect a (group-specific) normalization that prevents difficulties arising from groups being on different scales and is related to the scaled graphical lasso discussed in \citet{Stadler2013PenalizedModels}; see also \citet{Stadler2017MolecularStudy}.
The group-specific regularization parameters $\lambda_k$ in the graphical lasso are set by cross-validation. 
%In practice, we do so by setting $\lambda_k = \lambda \, \hat{n}_k^{-1/2}$, where $\lambda$ is  a single, global tuning parameter  that is set by CV. This ensures that group-specific regularization is in line with the scaling seen in universal penalization approaches \citep[see][ and references therein for details]{stadler2017} but requires CV for only a scalar, rather than $K$-dimensional, tuning parameter. 

In summary, \autoref{eq:glasso} allows  $K$ group-specific $p$-dimensional graphical models to be estimated by  $K$ calls to standard graphical lasso optimization. We note that essentially any graphical model estimator may be used in the final step as an alternative to the graphical lasso. Below we consider also the algorithm of \citet{Meinshausen2006High-dimensionalLasso} for this purpose.

\section{Results}
\label{sec:results}

We present empirical results investigating clustering performance in various settings, mostly where $p \! \gg \! n$.
We examine the performance of two versions of our proposed MCAP approach using PCA projections (\textit{MCAP-PCA} and \textit{MCAP-PCA-K}), where in the latter the projection dimension is fixed to the number of known groups $K$ in the data, as well as three versions based on random projections: \textit{MCAP-RP-Gauss} which uses a standard Normal projection matrix, the sparse variant \textit{MCAP-RP-Achl} and the very sparse variant \textit{MCAP-RP-Li}. 

We compare the proposed MCAP approach with the following range of existing clustering methods (labels used in figure legends are given here in parenthesis): the standard model-based clustering algorithm as implemented in the \texttt{mclust} package \citep{Scrucca2016MclustModels} using default settings (\textit{mclust}), the $\ell_1$-penalized mixture model included in the \texttt{nethet} \citep{Stadler2017MolecularStudy} Bioconductor package (\textit{MixGLasso}) plus an {\it ad hoc} variant that uses random subsampling of the variables (because the full penalized mixture model cannot be run on very large $p$ settings) such that $p_{\textnormal{max}}=500$ (\textit{MGL-sub500}), standard K-means as implemented in the R \texttt{stats} package (\textit{K-means}), hierarchical clustering using Euclidean distances and the Ward linkage option in the \texttt{hclust} package \citep{Murtagh2014WardsCriterion} (\textit{hclust}), and spectral clustering using the \texttt{kernlab} package \citep{Karatzoglou2004KernlabR} with a radial basis kernel (\textit{spectral}). 

We note that some methods, specifically \texttt{mixGlasso}, \texttt{mclust} and spectral clustering, were computationally infeasible to run (in some very large $p$ settings) or did not converge and therefore the results for these methods are absent in the corresponding figures.

\subsection{Overall strategy}
The empirical study of clustering performance is challenging due to the unsupervised nature of the problem. Using purely simulated data has the benefit that the true clusters are known and results can therefore be compared with a gold standard, but the disadvantage is that it is difficult to mimic the characteristics of real data. 
The latter point is particularly relevant to the present study, because (i) we emphasize the role of covariance structure  and (ii) realistic high-dimensional covariance or graphical model structure is hard to simulate. 

Motivated by these arguments, we use an approach  anchored in real data. In particular, we take  advantage of recent large-scale studies in biology (described below) where (i) groups in the data are clearly defined by  external physical aspects (in particular, cell type and cancer type) and (ii) total sample sizes are sufficiently large to allow extensive empirical study. 

In each case, the original data sets (the ``base data") $X$ consist of data matrices $X_1 \ldots X_K$ and the groups are known by design. We subsample the base data in various ways to test the ability to recover cluster assignments in different settings. 
Clustering methods are run on the subsampled data (with the true assignments always entirely hidden) and results are evaluated in a gold standard sense in terms of the adjusted Rand index with respect to the true labels. In all cases, the covariance structure is real as the data are subsamples of the real high-dimensional data.

We consider in particular varying sample size $n$, varying the dimension $p$, varying the magnitude of the mean signal $d$ and varying the number of groups $K$. 
Throughout this Section $n$ refers to the size of the sample actually available to the estimators in the experiment (as opposed to the total base data sample size); likewise $n_k$ refers to the corresponding group-specific sample size.
In examples below where the mean signal is stated as zero, this means that the group-specific data were centered (individually) to guarantee the absence of any mean signal.
%For the cancer data, where the total number of available groups is relatively large, we consider also varying $K$.
%We keep the focus on finite sample behavior, and all clustering methods were therefore applied to only the sampled data, without using the base data from which they were sampled. 

These real-data-based experiments have a certain differential covariance structure that is inherent to the problem. In order to study the effect of varying the magnitude of the covariance signal, we consider additional, purely simulated data. These include, in particular, the simple, K-means like case with {\it no} differential covariance structure. 

In order to focus the presentation, we do not discuss estimation of $K$ in this paper. This is a well-studied problem and, in particular for Gaussian mixture models, there are many approaches available \citep[see, for example,][]{Roeder1997PracticalNormals, Fraley2002Model-basedEstimation}. 
Our main focus here is on the feasibility of detecting the signals at all in challenging $p \gg n$ problems.
In the scenarios below, all methods are run with $K$ set to the known number of groups in the data.

\subsection{Single-cell data from neural cell types}
The data are from single-cell RNA-sequencing (scRNA-seq) assays \citep{Zeisel2014CellRNA-seq,Usoskin2015UnbiasedSequencing} of neural cells of two physically distinct types (dorsal root ganglia and hippocampal neurons) for which the true labels (cell type) are known.  
scRNA-seq has emerged in recent years as a widely used technology that allows gene expression to be quantified for individual cells as opposed to aggregates of many cells as in conventional RNA-seq or microarray assays.

The total dimension of the data is $p_{\mathrm{tot}}=10{,}280$ and the base data sample sizes are 864 and $1{,}314$ for the $K \eq 2$ cell types (dorsal root ganglia and hippocampal neurons), respectively. All results are based on subsampling these data and all clustering operations are carried out on the subsamples only and with no access to  the true labels.
In each case, the data were rank transformed to normality before clustering. This was done to  (i) remove ``easy" differences (e.g., in marginal variances) between the groups and (ii) to allow simple parameterization of the distance between groups (in order to consider behaviour for varying levels of mean signal).

\begin{figure}[ht]
	\centering    
    \includegraphics[width=0.8\textwidth]{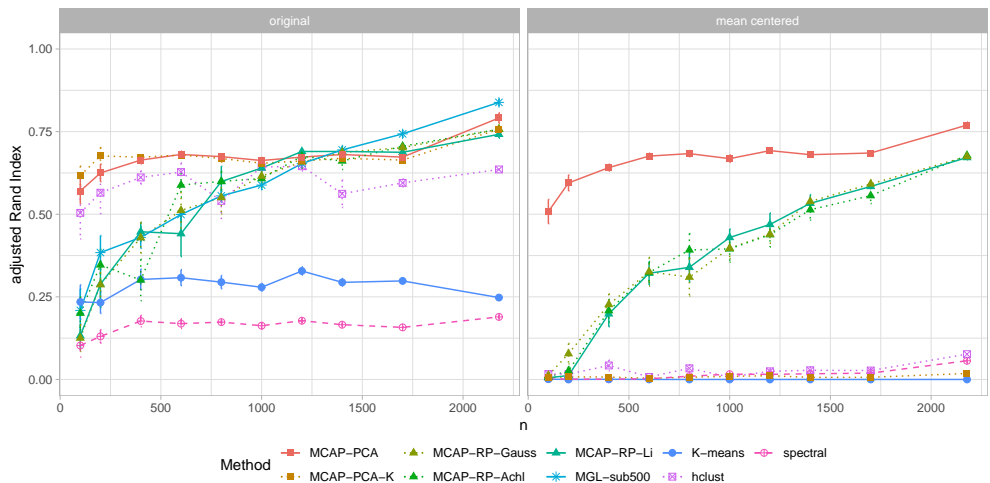}
    \caption[scRNA-seq results: varying $n$.]{Varying sample size $n$ on the single-cell RNA-sequencing data set ($K=2$). Adjusted Rand index as a function of total sample size $n$, with $n_k = n/2$ for each group and dimensionality $p_{\mathrm{tot}}$. \textit{Left}: Retaining the original mean signal per group. \textit{Right}: With the mean signal removed (that is, each group centered by design). Errorbars indicate the standard error of the mean over 10 realizations (data subsamples).}
    \label{fig:SC_vary_n}
\end{figure}

\begin{figure}[ht]
	\centering    
    \includegraphics[width=0.8\textwidth]{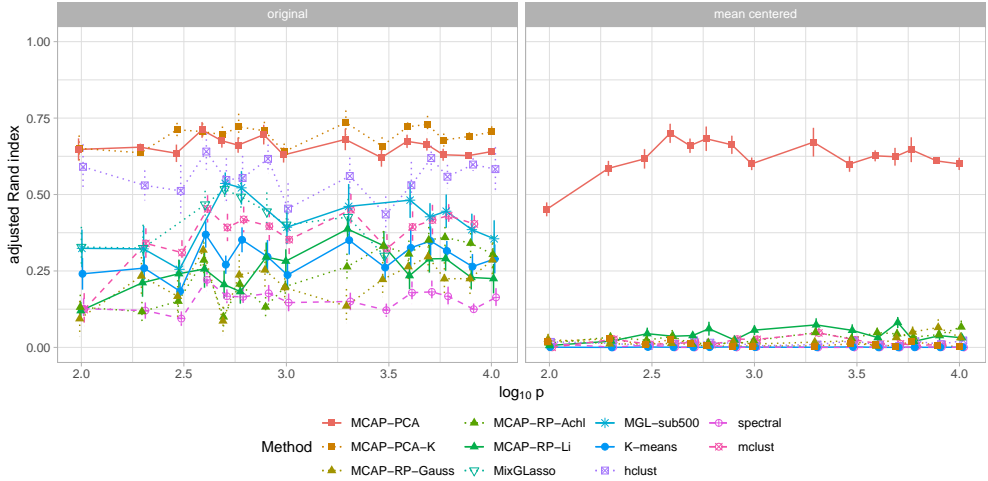}
    \caption[scRNA-seq results: varying $p$.]{Varying dimension $p$ on the single-cell RNA-sequencing set ($K=2$). Adjusted Rand index as a function of the number of variables $p$ (shown on a $\log$-scale) with $n_k$ fixed at 100 for each group. \textit{Left}: Retaining the original mean signal per group. \textit{Right}: With the mean signal removed. Errorbars indicate the standard error of the mean over 10 realizations.}
    \label{fig:SC_vary_p}
\end{figure}

\autoref{fig:SC_vary_n} shows adjusted Rand index as a function of total sample size $n$ ($n_k=n/2$ for each group; we consider non-identical group-specific sample sizes in the cancer example below), with $p$ fixed to $p_{\mathrm{tot}}$. The panel on the right in \autoref{fig:SC_vary_n} shows the same regimes with the mean signal entirely removed. 
\autoref{fig:SC_vary_p} shows adjusted Rand index as a function of $p$, with $n_k$ fixed to 100 in each of the two groups.
Again, the right panel has the mean signal  set to zero. Here, for a given dimension $p$, data were  obtained by randomly subsampling the indicated number $p \leq p_{\mathrm{tot}}$ of variables from the complete set. 
%Results shown are the average over 10 subsampled realizations for each point in the graph with errorbars indicating the standard error of the mean. 

It is interesting to compare the performance of projection down to $K$ (in this case 2) dimensions (\textit{MCAP-PCA-K}) with the adaptive strategy. When  the mean signal is present (left hand panels), the two-dimensional projection is already sufficient to recover assignments. However, when the mean signal is removed, leaving only the covariance signal, only the adaptive projection is effective. 
This provides a real-data example of the bias-variance-type tradeoff described in \autoref{sec:risk}: When the target dimension $q$ is too low, the covariance signals in the high-dimensional space are lost under projection (even as sample size increases, as seen for \textit{MCAP-PCA-K} in the right panel in \autoref{fig:SC_vary_n}; but, these signals can be detected with an appropriate choice of $q$.

\begin{figure}[ht]
	\centering    
    \includegraphics[width=0.65\textwidth]{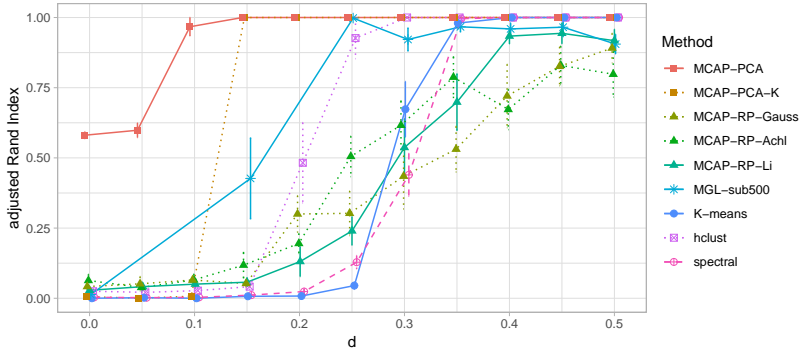}
    \caption[scRNA-seq results: varying $d$.]{Varying mean signal strength $d$ on the single-cell RNA-sequencing data set ($K=2$). Adjusted Rand index as a function of the mean signal parameter $d$ (see text for details on how the group means are shifted according to $d$), with $n_k$ fixed at 100 for each group and $p = p_{\mathrm{tot}}$. Errorbars indicate the standard error of the mean over 10 realizations (data subsamples).}
    \label{fig:SC_vary_d}
\end{figure}

\autoref{fig:SC_vary_d} shows adjusted Rand index as a function of the mean signal parameter $d$, with $n_k$ fixed to 100 in each of the two groups and $p$ fixed to $p_{\mathrm{tot}}$. 
The mean signal was set using the value $d$ shown in the plot as follows: after centering the data from each group, one of the groups was shifted by a random sign vector multiplied by the scalar $d$. Thus, larger $d$ corresponds to a larger mean signal. %Results are shown over 10 datasets. 
Here, we see a phase transition-like behaviour for most of the clustering methods with performance rapidly improving above a certain magnitude of mean signal.
However, \textit{MCAP-PCA} is notably more effective than any of the other methods in the sense that it is the only approach that can detect the pure covariance signal (i.e.\ in the $d=0$ case); furthermore, as $d$ increases, it achieves near perfect performance before the other methods even start to improve. 
For completeness, we show in \autoref{fig:sim_vary_d} (\hyperref[{app:sim_kmeans}]{Appendix A}) an example with $d$ varying for data simulated from isotropic Gaussians. This represents a simpler case that better fits the mean-based methods---as expected, K-means performs well, but is matched by MCAP. 

\begin{figure}[ht]
	\centering    
    \includegraphics[width=0.8\textwidth]{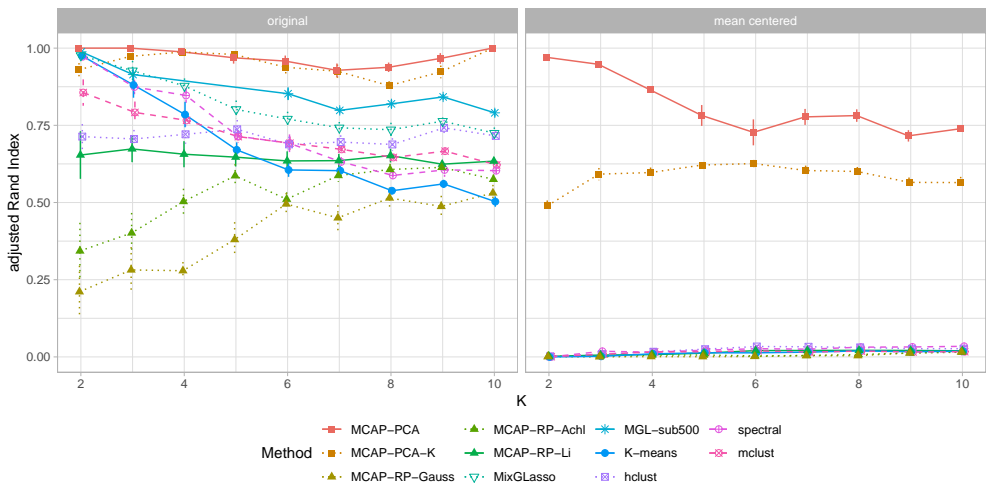}
    \caption[scRNA-seq results: varying $K$.]{Varying the number of groups $K$, based on the single-cell RNA-sequencing data set. Adjusted Rand index as a function of $K$ (see text for details on how groups for $K>2$ are simulated using the original data). The number of samples per group is $n_k=100$ and the dimension is fixed to $p=1{,}000$. \textit{Left}: Retaining the original mean signal per group. \textit{Right}: With the mean signal removed. Errorbars indicate the standard error of the mean over 10 realizations (data subsamples).}
    \label{fig:SC_vary_k}
\end{figure}

\autoref{fig:SC_vary_k} shows adjusted Rand index as a function of the number of groups $K$ for $n_k=100$ and $p=1{,}000$. In order to simulate $K>2$, only one group (dorsal root ganglia) was used from the original data, and different covariance signals were induced by randomly sampling $p=1{,}000$ different variables (genes) for each group. 
%More precisely, for each group in each of the scenarios for $K$ in \autoref{fig:SC_vary_k}, $p<p_{\mathrm{tot}}$ variables were subsampled at random without replacement. 
This approach ensures non-identical covariance signals for all $K$ groups that are realistic in the sense that all variables and correlations are real. 
However, the groups created in this way, while based on real data, do not represent scientifically meaningful clusters; we consider an example with $K>2$ scientifically meaningful groups (cancer types) below.
In \autoref{fig:SC_vary_k} we see that the good performance of \textit{MCAP-PCA} is maintained as the number of clusters increases, demonstrating that its behaviour is not a special case for a specific $K$.

%\FloatBarrier

In our approach, a proxy for the assignment risk $R$ is used to set the target dimension. However, due to the unsupervised nature of the mixture modelling problem, it is difficult to precisely estimate assignment risk $R$ and  therefore instructive to consider sensitivity to the target dimension $q$. In the present real data example, the true assignments are known by design, hence we can empirically investigate the sensitivity of assignment risk to the target dimension and consider the estimated target dimension $\hat{q}$ in light of a corresponding oracle. 

\begin{figure}[t]
	\centering    
    \includegraphics[width=1.0\textwidth]{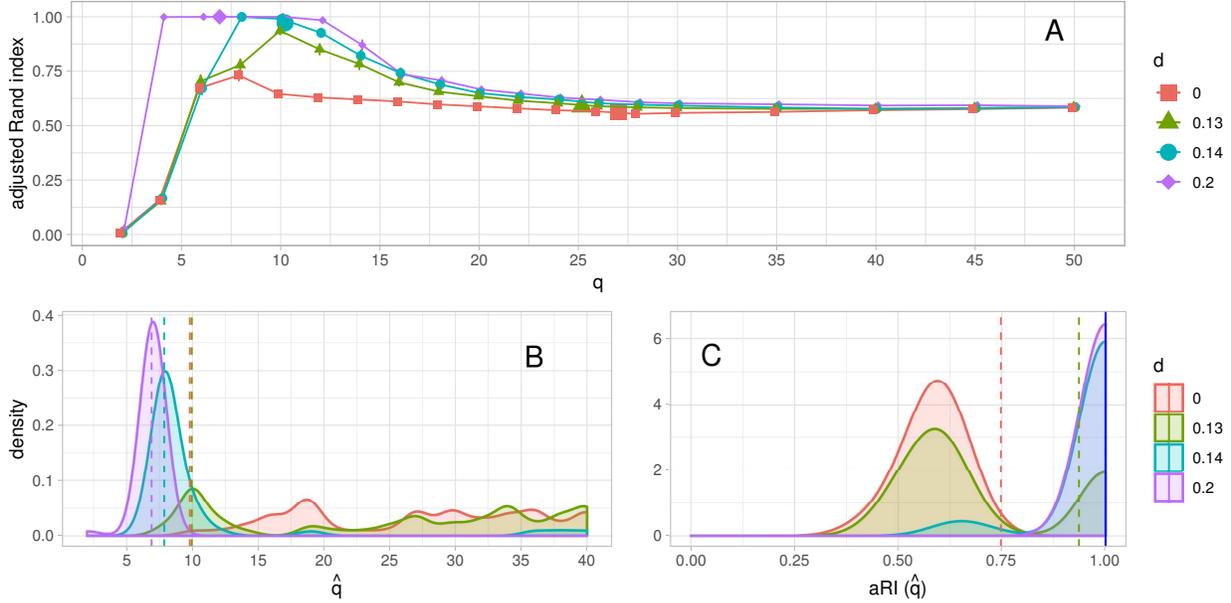}
    \caption[scRNA-seq results: Adjusted Rand index as a function of $q$.]{Sensitivity to target dimension $q$. Results are based on the single-cell RNA-sequencing data set for four different mean signals $d$ and parameters $K=2$, $n=200$ ($n_k=100$), $p=p_{\text{tot}}$; based on 50 realizations (data subsamples). \textbf{(A)} Adjusted Rand index (aRI) as a function of $q$. Bold symbols indicate the average optimal target dimension $\hat{q}$ as determined by \textit{MCAP-PCA}. \textbf{(B)} Kernel density estimates (Gaussian, bandwidth of 1) of $\hat{q}$. Dashed lines indicate the true (oracle) optimum $q^*$ (as given by the highest aRI). \textbf{(C)} Kernel density estimates (Gaussian, bandwidth of 0.06) of adjusted Rand index corresponding to the data-adaptive $\hat{q}$'s in panel B. Dotted vertical lines indicate performance using the oracle values $q^*$. (Note that for $d=0.14$ and $d=0.20$, the oracle values coincide at 1.00.)}
    \label{fig:SC_vary_q}
\end{figure}

Panel A in \autoref{fig:SC_vary_q} shows Rand index (higher values indicate lower assignment risk) as a function of target dimension $q$ for the scRNA-seq data (with \mbox{$p=p_{\mathrm{tot}}$}, $K=2, n_k=100$) for four different levels of mean signal $d$. We see that for a higher mean difference the optimal target dimension $q$ decreases. This provides a real data example of the phenomenon described in \autoref{sec:risk}, where, as the mean difference increases, the signal becomes  likely to be retained in the first few principal components. Note that we only show $q$ up to 50, hence the large increase in variance with larger $q$ is not shown. For each curve, the mean over 50 realizations of the automatically set projection dimension $\hat{q}$ is shown as a bold symbol. 

\autoref{fig:SC_vary_q}B,C show distributions over estimated values $\hat{q}$ and the corresponding Rand indices achieved by \textit{MCAP-PCA}. The values $q^*$ that are optimal in a label oracle sense (that is, resulting in the highest Rand index) are shown as vertical dashed lines. We see that MCAP can detect the need for a smaller $q$ as the mean signal increases and that the Rand index performance is typically close to the oracle. In the challenging $d=0$ case, $\hat{q}$ is more variable and typically larger than $q^*$. However, the corresponding Rand indices (Panel C) show that this does not translate to high variability or poor performance in the assignment risk sense; the results are in fact fairly consistent and typically a little worse than the oracle. This is due to the fact that in this case the sensitivity to $q$ is low (as also seen in Panel A) and for this reason there is little signal to guide the setting of $q$. This illustrates an appealing feature of stability-based setting of $q$---although it  is a rough proxy, it is more effective in the setting where it is most needed, i.e.\ when there is a larger effect on assignment risk or sensitivity with respect to $q$, as exemplified by the larger $d$ examples here.

\FloatBarrier

\subsection{Data from cancer samples}
The data originate from The Cancer Genome Atlas (TCGA, \mbox{\url{http://cancergenome.nih.gov/}}), a large-scale cancer patient study. We obtained the (batch) RNA-seq data from the TCGA database and extracted RSEM \citep{Li2011RSEM:Genome} estimates of the gene frequencies as a measure of gene expression. These gene expression values were further rank-normalized in the same way as the single-cell RNA-seq data. 
 
For our application, the base data consist of the gene expression data from the four largest sample size cancer types in TCGA, namely  breast ({\it BRCA}), kidney renal clear cell ({\it KIRC}), lung adenocarcinoma ({\it LUAD}) and thyroid ({\it THCA}).
Thus, each sample comes from a distinct cancer type which is used as the true cluster label.
The total dimension of the data is $p_{\mathrm{tot}}=20{,}531$ and the total sample sizes for the different cancer types are 1,216 ({\it BRCA}), 606 ({\it KIRC}), 576 ({\it LUAD}) and 572 ({\it THCA}), respectively. As with the first data set, all results are based on subsampling these data and all clustering operations are carried out on the subsamples only and with no access to the true cancer type labels.

\begin{figure}[ht]
	\centering    
    \includegraphics[width=0.8\textwidth]{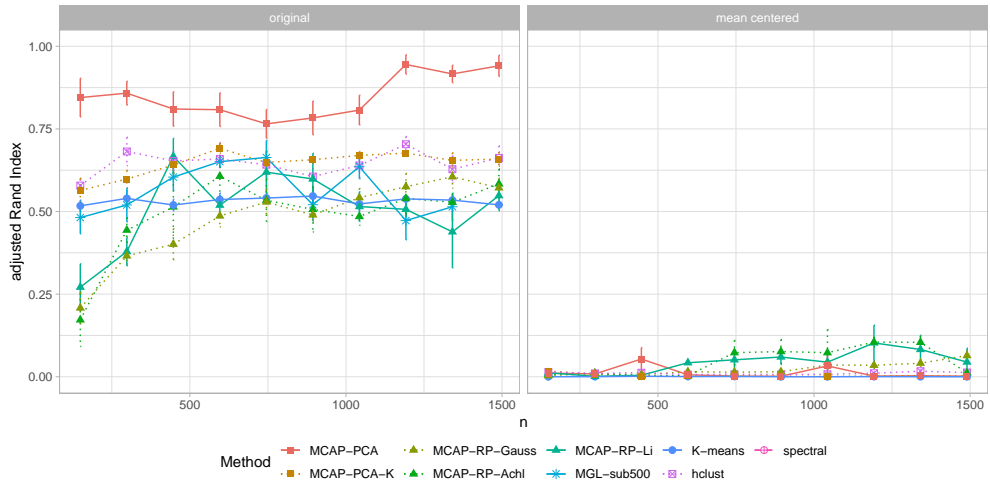}
    \caption[TCGA results: varying $n$.]{Varying sample size $n$ on the cancer data set ($K=4$). Adjusted Rand index as a function of total sample size $n$ (the maximum $n_k$'s are 1,216 ({\it BRCA}), 606 ({\it KIRC}), 576 ({\it LUAD}) and 572 ({\it THCA})); $p = p_{\mathrm{tot}}$. \textit{Left}: Retaining the original mean signal per group. \textit{Right}: With the mean signal removed (that is, each group centered). Errorbars indicate the standard error of the mean over 10 realizations (random data subsamples).}
    \label{fig:TCGA_vary_n}
\end{figure}

\autoref{fig:TCGA_vary_n} shows adjusted Rand index as a function of $n$, with $p$ fixed to $p_{\mathrm{tot}}$. The $n$ samples were subsampled at random from the base data, giving non-identical group-specific sample sizes with relative sample sizes in line with the base data.  
%Results are shown over 10 realisations for each data point in the figure. 

\begin{figure}[ht]
	\centering    
    \includegraphics[width=0.8\textwidth]{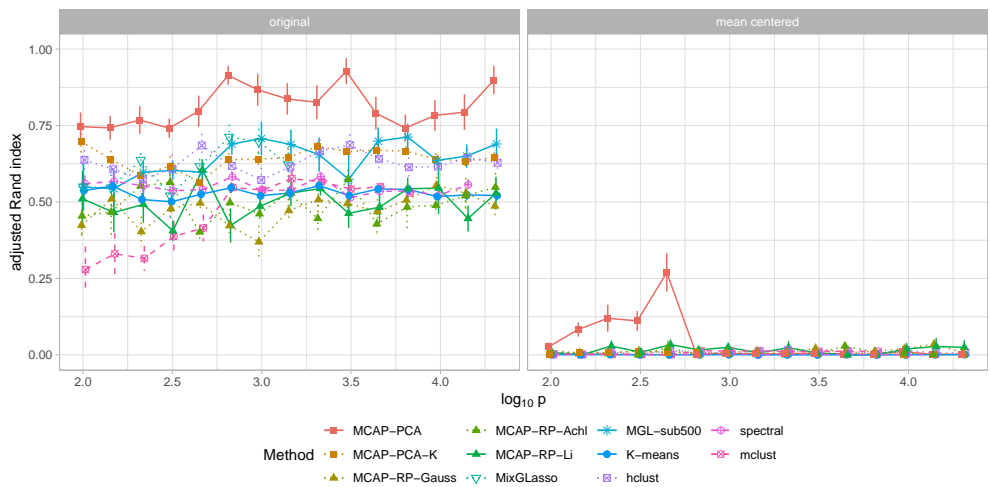}
    \caption[TCGA results: varying $p$.]{Varying dimension $p$ on the cancer data set ($K=4$). Adjusted Rand index as a function of the number of variables $p$ (shown on a $\log$-scale), with number of samples per group fixed to $n_k=100$. \textit{Left}: Retaining the original mean signal per group. \textit{Right}: With the mean signal removed. Errorbars indicate the standard error of the mean over 10 realizations (data subsamples).}
    \label{fig:TCGA_vary_p}
\end{figure}

\autoref{fig:TCGA_vary_p} shows adjusted Rand index as a function of $p$, with $n$ fixed to 593 ($=0.2n_{\mathrm{tot}}$). Again, the $n$ samples were subsampled at random from the base data, giving non-identical group-specific sample sizes with relative samples sizes in line with the base data (resulting in $n_k$'s of 243, 121, 115 and 114 for {\it BRCA}, {\it KIRC},  {\it LUAD} and {\it THCA}, respectively). The dimension was varied by randomly subsampling the indicated number $p \leq p_{\mathrm{tot}}$ of variables from the complete set. 
MCAP is again highly effective in this topical example from cancer biology and clearly outperforms the other methods. 
However, with these data, $d=0$ provides a failure case where none of the methods, including MCAP, are able to detect the differential signal.

\begin{figure}[ht]
	\centering    
    \includegraphics[width=0.65\textwidth]{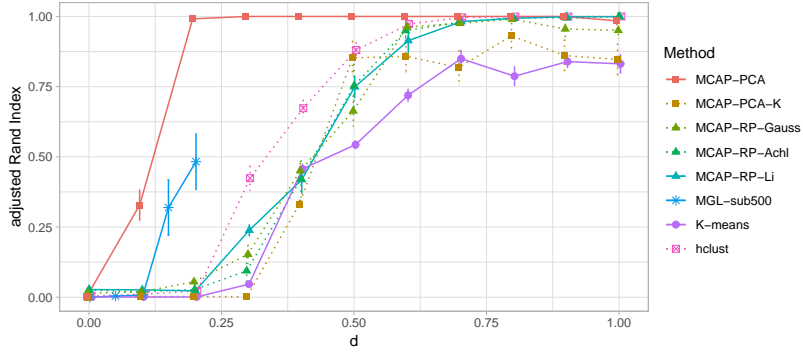}
    \caption[TCGA results: varying $d$.]{Varying strength of mean signal $d$ on the cancer data set ($K=4$).  Adjusted Rand index as a function of the mean signal parameter $d$ (see text for details on how the group means are shifted according to $d$), with number of samples per group fixed to $n_k=100$ and $p = p_{\mathrm{tot}}$. Errorbars indicate the standard error of the mean over 10  realizations (data subsamples).}
    \label{fig:TCGA_vary_d}
\end{figure}

\autoref{fig:TCGA_vary_d} shows adjusted Rand index as a function of $d$, with $n$ fixed to 593 
($n_k$'s were 243, 121, 115 and 114, as above) and $p=p_{\mathrm{tot}}$. The mean signal was controlled via the parameter $d$ as in \autoref{fig:SC_vary_d}. Since the number of groups is greater than 2, this was done by fixing one group as the ``centre", and then shifting each of the remaining groups by a random sign vector multiplied by the scalar $d$.
Again, larger $d$ corresponds to a larger mean signal (albeit in a more complicated way than in the $K=2$ case). 
In this second real data example with $K=4$ we see again that, as $d$ increases, \textit{MCAP-PCA} reaches near perfect performance much earlier than any of the other methods.

\begin{figure}[ht]
	\centering    
    \includegraphics[width=0.8\textwidth]{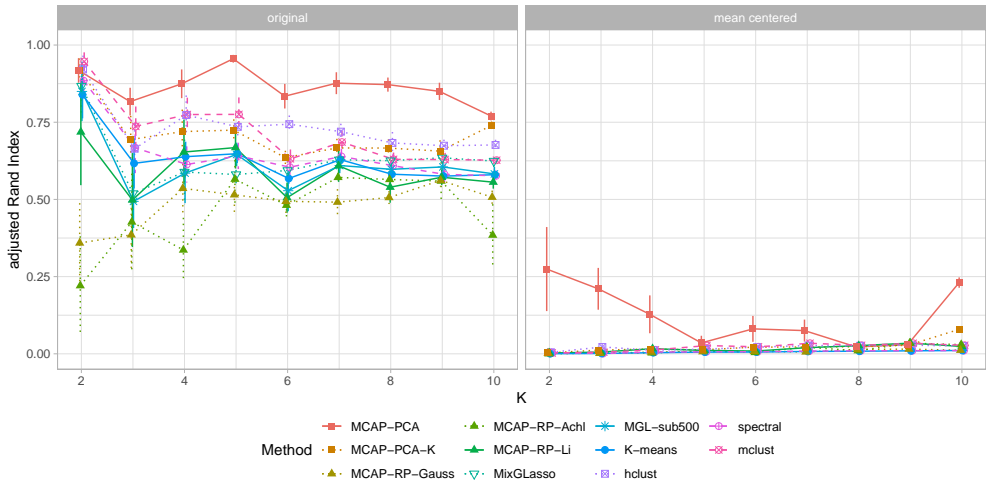}
    \caption[TCGA results: varying $K$.]{Varying the number of groups $K$ based on the extended cancer data set (including 10 distinct cancer types). Adjusted Rand index as a function of $K$. The number of samples per group is fixed to $n_k=100$ and dimension to $p=1,000$. \textit{Left}: Retaining the original mean signal per group. \textit{Right}: With the mean signal removed. Errorbars indicate the standard error of the mean over 10 realizations (data subsamples).}
    \label{fig:TCGA_vary_k}
\end{figure}

\autoref{fig:TCGA_vary_k} shows adjusted Rand index as a function of $K$, with $n_k$ fixed to 100 and $p$ fixed to $1{,}000$. Here, in contrast to the previous varying $K$ example (\autoref{fig:SC_vary_k}), the groups  are scientifically meaningful, representing different cancer types. 
We reduced $p$ to allow comparison with the penalized mixture models. However, we note that MCAP can be easily run on much higher-dimensional data, as shown in many of the other examples.

\subsection{High-dimensional graphical model estimation}

In the examples above we focused on assignment risk in high-dimensional mixture modelling but did not directly discuss estimation of the high-dimensional parameters themselves. In some settings, it is the latter and not just the assignments, that are of interest. For example, the group-specific graphical models parameterized by $\Omega_k$ are often of direct scientific interest. In this section we show results concerning recovery of these high-dimensional parameters. 

The experiments in the previous sections were designed to test the ability of MCAP under conditions of nontrivial real-world high-dimensional covariance structure. However, these data are not ideal for testing recovery of graphical models because the true data-generating parameters or graphs are not known. 
%These could be estimated from the data, but this would bias the analysis, since if the  estimator being evaluated  uses similar assumptions to the estimator used to define the ground truth, they are more likely to agree. 
Therefore, in the following we use fully  simulated data generated from known graphical models. This allows us to directly quantify the ability of the estimation strategies described in \autoref{sec:param_estimation} to recover the high-dimensional graphical model structure.

The simulation strategy was as follows. Data were generated from $K=2$ Gaussian graphical models with (known) undirected graphs $G_1$ and $G_2$. The graphs consisted of randomly located edges and the data were generated using the \texttt{R} package \texttt{huge} (specifically the function \texttt{huge.generate}).
As usual, the group labels were hidden and the data matrix $X$ provided as input to mixture modelling. This gave rise to estimated group assignments as well as estimated parameters, including inverse covariance matrices $\hat{\Omega}_k$ and corresponding graphical models $\hat{G}_k$.
To allow direct comparison with \texttt{mixGLasso}, we restricted the dimension to $p=500$ (with approximately 100 edges in each graph $G_k$) but note that MCAP can be run on much higher dimensional examples (see below).

\begin{figure}[ht]
	\centering    
    \includegraphics[width=1.0\textwidth]{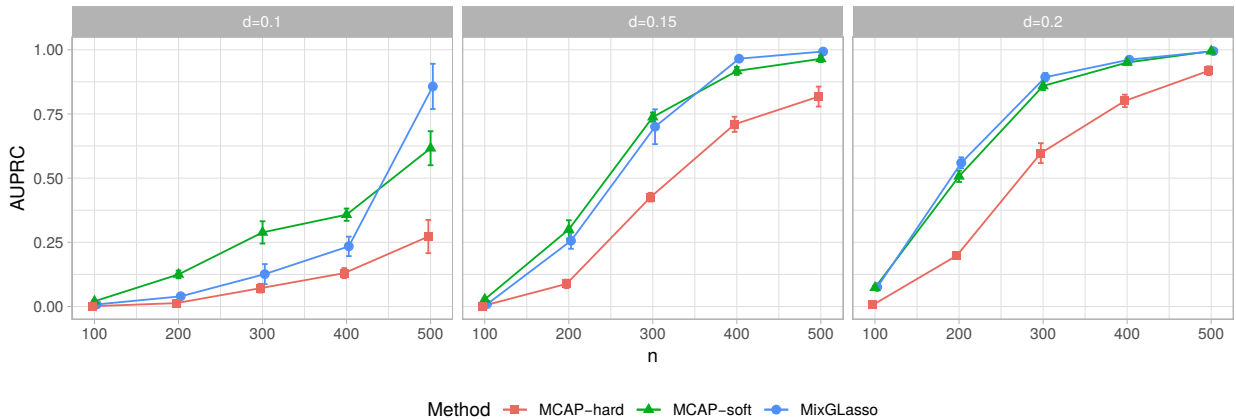}
    \caption[Simulation results: AUPRC as a function of $n$ and different $d$.]{Simulated data to investigate graphical model recovery. The x-axis indicates different settings of total sample size $n$. The y-axis shows the area under the precision-recall curve for true edge recovery after ``hard" and ``soft" graphical model recovery following MCAP (see text for details). Results are based on simulated data with $K=2$, $n_k = n/2$, $p=500$ and number of edges in the true graphs roughly $100$ (see text for further details). The three panels show results for three different levels of mean signal (indicated over each panel). Errorbars indicate the standard error of the mean over 10 realizations.}
    \label{fig:sim_aucpr}
\end{figure}

\autoref{fig:sim_hits} shows the fraction of edges recovered in the estimates $\hat{G}_k$, specifically the fraction of the true positives among the top 100. This corresponds to a strict threshold in an ROC sense, since only 100 (out of $p^2 \eq  250{,}000$) edges are considered. \autoref{fig:sim_aucpr} shows the area under the precision-recall curve for the recovery of true edges after clustering. Edge recovery is effective and as expected improves with $n$ or increasing mean signal. Weighted estimation outperforms hard assignment and is competitive with \texttt{mixGlasso} (but at much lower computational cost and better scalability). We note that model recovery did not work in the $d=0$ setting. 

To test the feasibility of using MCAP for parameter estimation in the larger $p$ setting using real data, we ran \textit{MCAP-PCA} on the single-cell RNA-seq data, allowing $p$ to vary following the sampling approach used above (as in Figure \ref{fig:SC_vary_p}). Using Meinshausen-B\"{u}hlmann estimation (with hard assignments), MCAP is highly scalable and can be rapidly run on the full dimensional problem ($p= 10,280$) using standard desktop resources (e.g., this took approximately 90 minutes of serial compute time on a standard workstation). Note that these experiments on the RNA-seq data were performed to consider computational feasibility; we do not know the true underlying graphical model for the data and did not attempt to assess the resulting estimates. We were not able to compare with \texttt{mixGlasso} in this case due to computational limitations.

\vspace{0pt}

%\FloatBarrier

\begin{figure}[ht]
	\centering    
    \includegraphics[width=1.0\textwidth]{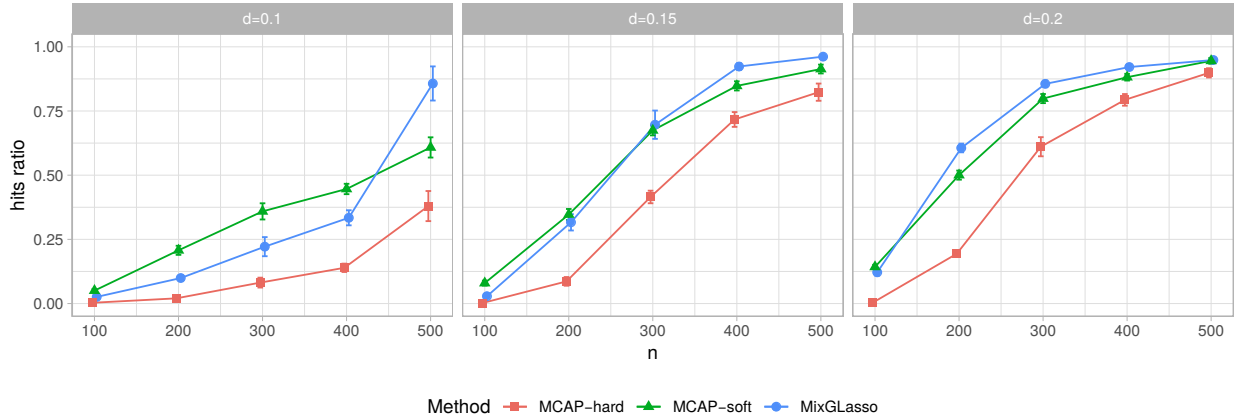}
    \caption[Simulation results: graphical model recovery.]{Simulated data to investigate graphical model recovery. The x-axis indicates different settings of total sample size $n$. The y-axis shows the fraction of true edges recovered after ``hard" and ``soft" graphical model recovery following MCAP (see text for details). Results based on simulated data with $K=2$, $n_k = n/2$, $p=500$ and number of edges in the true graphs roughly $100$ (see text for further details). The three panels show results for three different levels of mean signal $d$ (indicated over each panel). Errorbars indicate the standard error of the mean over 10 realizations.}
    \label{fig:sim_hits}
\end{figure}

\FloatBarrier

\subsection{Additional simulation results varying the covariance signal, including the mean-signal-only case}

The real-data-based experiments in the previous sections have a certain level of differential covariance structure inherent to the data. In order to consider the effect of systematically varying the covariance signal, in this section we present additional experiments using purely simulated data. 
In brief, we use a block diagonal set-up, with block covariances drawn from an inverse Wishart with specified degrees of freedom $w$. The latter allows control of the covariance signal: when $w$ is small, the covariance signal is larger and when $w$ is big, the covariance signal (asymptotically) disappears. 

Specifically, for each group $k = 1,2$ we sample a $B$-dimensional covariance matrix $\Sigma_k^{(B)} \sim \mathrm{IW}(w,A)$, where $\mathrm{IW}(w,A)$ denotes an inverse Wishart distribution with degrees of freedom $w$ and scale matrix $A$. The parameter $B$ is set to $5{,}000$ throughout, $A$ to the identity matrix $\textnormal{I}_B$ and $w \in \{B, 2B, \infty\}$.
Then, for a specified total dimension $p$, the group-specific data are drawn from a multivariate Normal distribution (MVN) with a block-diagonal covariance matrix where each block is identical to $\Sigma_k^{(B)}$. In the case of $w=\infty$, there is no covariance signal. (In practice, for this case we simply use the identity matrix to sample form the MVN instead of drawing from the inverse Wishart first). 

We consider three different settings for the mean signal $d \in \{0, 0.05, 0.1\}$, realized as a randomly signed shift per variable for the second group (as described earlier). This gives a total of nine regimes (three levels of covariance signal times three levels of mean signal).

\begin{sidewaysfigure}[ht]
	\centering    
    \includegraphics[width=1.0\textwidth]{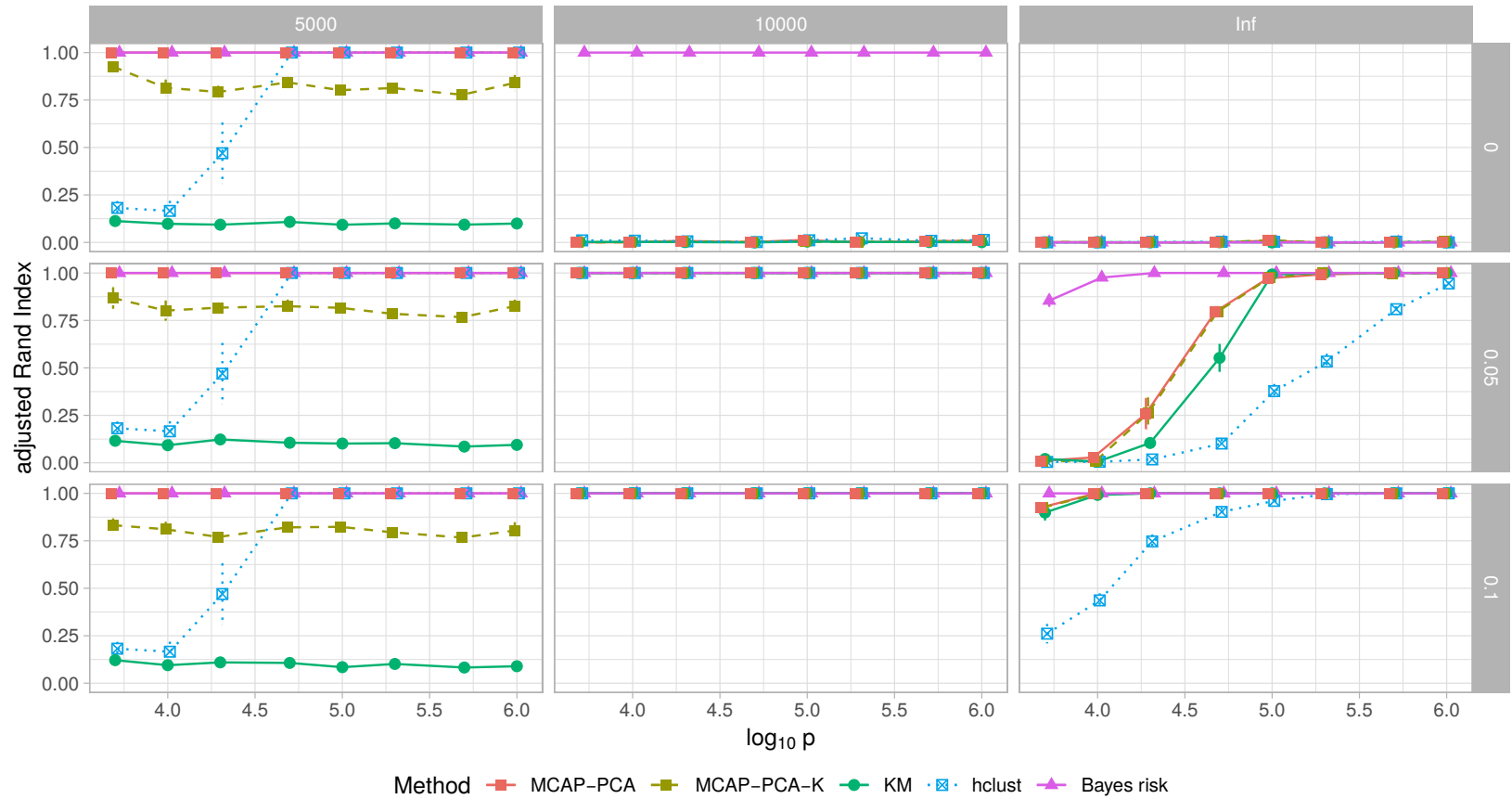}
    \caption[Simulation results: varying $p$ for different settings of Wishart $dof$ and $d$.]{Simulated data to assess performance when varying the covariance signal (see text for further details on how the data was generated). The y-axis shows dimensionality $p$ on a $\log_{10}$-scale. Columns show different settings of the Wishart degrees of freedom (indicated at the top) and rows show three scenarios for different mean signal $d$ (indicated on the side);  with $K=2$, $n_k = 100$. Errorbars indicate the standard error of the mean over 10 realizations.}
    \label{fig:sim_vary_sigma_full}
\end{sidewaysfigure}

\autoref{fig:sim_vary_sigma_full} shows results for these nine scenarios with column-wise decreasing covariance signal and row-wise increasing mean signal. Shown is performance (as measured by the adjusted Rand index) of the \textit{MCAP-PCA} and \textit{MCAP-PCA-K} variants in comparison with K-means and hierarchical clustering as a function of $p$ up to a dimensionality of $10^6$. Additionally, the Bayes risk is given to indicate the best possible results when the true data-generating distributions are known (that is, the classification performance using the true data-generating distributions). 
%This was obtained by classifying each sample into the class with the higher group-specific probability density at that data point...

Since sampling very high dimensional data from inverse Wishart and multivariate Normal distributions is computationally demanding, we also considered a simpler simulation setup where we sample base data sets of $p^{(0)} = 10^4$, i.e.\ two identical blocks of size $B=5{,}000$ for the block-diagonal covariance matrix, and base data matrices $X_k^{(0)} \sim MVN$ of size $n_k^{(0)}=10^4$ samples. In each experiment, the actual data matrices $X_k$ are subsampled using $X_k^{(0)}$. Data matrices for scenarios with $p>p^{(0)}$ were created by concatenating $n_k$ samples of the base data with additional, randomly subsampled entries of randomly chosen columns of the base data matrix belonging to the given group. Results are shown in \autoref{fig:sim_vary_sigma} (\hyperref[{app:sim_sigma}]{Appendix B}).

In these and the additional experiments in \hyperref[{app:sim_kmeans}]{Appendix A}, it is interesting to compare MCAP to K-means in the setting where there is little or no covariance signal. These experiments show that the cost of allowing for covariance  structure via MCAP is small, in the sense that even when the covariance signal is entirely absent, MCAP performs as well as, or better than, the classical mean-based methods.

%\FloatBarrier

\section{Discussion}
\label{sec:discussion}

We proposed an approach called MCAP by which to perform model-based clustering in very high dimensions, whilst accounting for both mean and covariance signals. The key idea is to combine projections---where the target dimension is set in a data-adaptive manner---with full covariance mixture modelling. This allows detection of both types of signal whilst controlling computational and statistical cost. We showed that the proposed approach is effective in a range of high-dimensional examples, spanning  combinations of sample sizes, number of clusters and the type of signal present. 

\newpage 

\FloatBarrier

Two interesting theoretical aspects with respect to linear projections that are beyond the scope of this paper are (i) to understand why PCA is effective in the settings considered here and (ii) to better understand the properties of random projections for clustering with covariance signals.
We found MCAP (with PCA as the projection) to be highly effective, even in very high dimensions.
There are a number of insights from random matrix theory concerning PCA in the high-dimensional setting that point to potential difficulties, essentially due to fluctuations in the spectrum of the sample covariance matrix that arise even in the ``null" case of no low-dimensional structure \citep{Johnstone2001OnAnalysis}.
However, in our setting, the relevant issue  is not consistency of the PCA model per se, nor subspace recovery \citep{Jung2009PCAContext}, but rather the weaker requirement that signals that separate the groups are retained in the low-dimensional space. 
Here there are also connections to recent work on discriminative/generative learning and differential estimation in high dimensions, as discussed, for example, in \citet{Prasad2017OnModels}.

% * <fdondelinger.work@gmail.com> 2018-09-07T13:27:56.532Z:
% 
% I think currently the major weakness of the paper is that we don't have a satisfying answer to the question: why does PCA outperform random projects? Not sure if there's anything we can do about it, but perhaps it needs some more thought?
% 
% ^.
%Recent work on high-dimensional quadratic discriminant analysis provides a useful perspective \citep{jiang2015quda}.

For random projections, the Johnson-Lindenstrauss lemma \citep{Dasgupta2003AnLindenstrauss} concerns the behaviour of relative distances under projection and offers key theoretical support for random-projection-based clustering. 
Indeed, we found that in our experiments with increasing mean signal (\autoref{fig:SC_vary_d} and \autoref{fig:TCGA_vary_d}), variants of MCAP based on random projections performed reasonably well for sufficiently large mean signals.
On the other hand, for covariance signals, random projection was not effective and even in the increasing mean signal case, PCA-based MCAP was always more effective, often by a wide margin.
However, a  variety of random projections beyond the classical i.i.d.\ Gaussian approach or the sparse variants that we used here have been studied in the literature and it is possible that another choice would give better results. For example, \citet{Cannings2017Random-projectionClassification} have recently studied the use of {\it ensembles} of random projection realizations for classification. Similarly for unsupervised learning, it may be fruitful to consider multiple realizations of the projection matrix itself, selecting or up-weighting those that show evidence of retaining relevant signals.

In supervised learning, there is a well-known connection between regression following PCA (``principal components regression" or PCR) and $\ell_2$-regularized or ridge regression, namely that the latter shrinks low-variance directions more strongly, while the former entirely discards them \citep[see][for a lucid account]{Hastie2009TheLearning}. 
As in PCR, the regularization in MCAP is due to the retention of only a few PCs for the second stage estimation. This view of the role of the target dimension $q$ provides another perspective on the bias-variance tradeoff that motivated our data-adaptive strategy.

%We did not consider in detail parameter estimation in the original space. Here, a key point is that if  cluster assignments are sufficiently good, the problem reduces to a conventional  high-dimensional estimation task, either for each group separately or jointly as in \citet{Danaher2014TheClasses}. 
%In either case, the essential point is that by dealing with the latent indicators in the low-dimensional space brings considerable computational and statistical gains.
The statistical efficiency of mixture modelling in very high dimensions---that is, without a projection step---remains unclear. Although several penalized schemes have been proposed \citep{Zhou2009PenalizedMatrices,Stadler2017MolecularStudy}, in practice the detailed formulations, net of tuning parameters, etc., may not yet be optimal. Hence, the fact that MCAP in some cases outperformed an exemplar of this type of approach (\texttt{mixGlasso}) does not imply that mixture modelling directly in the high-dimensional space is infeasible, but rather points to the need for more work on both theory and methodology for penalized mixtures in the large-$p$ setting.

%%%%%%%%%%%%%%%%%%%%%%
\section*{Acknowledgements}
\addcontentsline{toc}{section}{Acknowledgements}
We would like to thank Richard Samworth and Sara van de Geer for useful discussions.

\section*{Code}
Source code (\texttt{R} package) for MCAP is available at \mbox{\url{https://github.com/btaschler/mcap}} .

\FloatBarrier

\bibliographystyle{abbrvnat}
\addcontentsline{toc}{section}{References}
\bibliography{references}

\appendix
\section*{Appendix A}
\label{app:sim_kmeans}

We present some additional experiments concerning the very large-$p$ case and simulations from a K-means-like isotropic  Gaussian model, i.e.\ with no covariance signal.

\begin{figure}[ht]
	\centering    
    \includegraphics[width=1.0\textwidth]{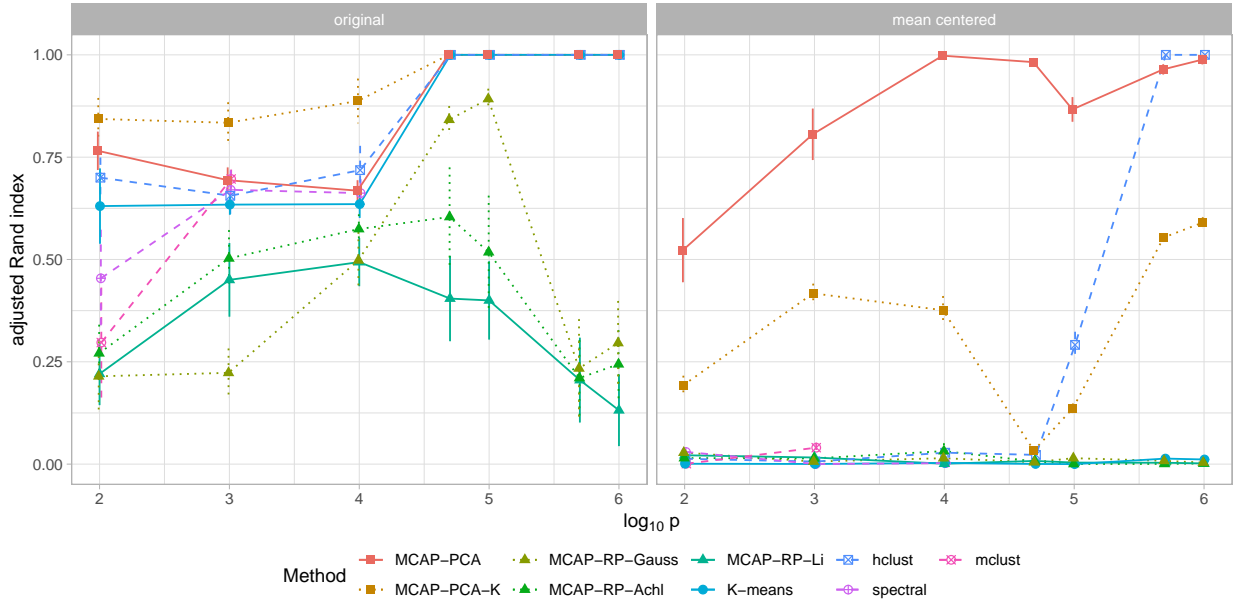}
    \caption[scRNA-seq results: very large $p$.]{Very large $p$ example based on single-cell RNA-sequencing data set ($K=2$). Adjusted Rand Index as a function of dimensionality $p$ (shown on a $\log$-scale; see text for details on how the data was created) with $n_k=100$ samples per group. \textit{Left}: Retaining the original mean signal per group. \textit{Right}: With the mean signal removed. Errorbars indicate the standard error of the mean over 10 realizations.}
    \label{fig:SC_vary_large_p}
\end{figure}

\autoref{fig:SC_vary_large_p} shows adjusted Rand Index as a function of very large $p$, based on the single-cell RNA-seq data set with $K=2$ and $n_k=100$. The data sets were created by concatenating multiple random draws of $n_k$ samples, with a different permutation of all $p_{\mathrm{tot}}$ genes per draw, for each group. 
This creates data with differential covariance structure in very high dimensions, but where each variable is real. The overall covariance structure is, however, artificial since it is based on permutations of one dataset.

\begin{figure}[ht]
	\centering    
    \includegraphics[width=1.0\textwidth]{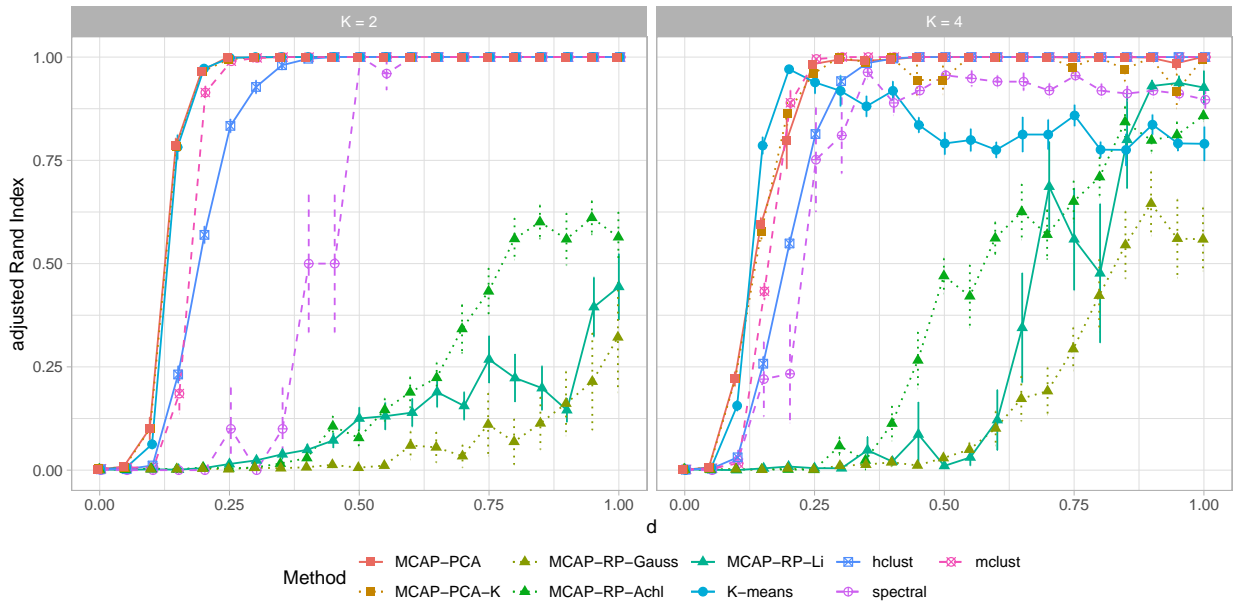}
    \caption[Simulation results: varying $d$.]{Varying mean signal strength $d$ based on simulated data from isotropic Gaussians with $n_k=100$ and $p=1000$. Adjusted Rand Index shown as a function of the mean signal parameter $d$. \textit{Left}: $K=2$. \textit{Right}: $K=4$. Errorbars indicate the standard error of the mean over 10 realizations.}
    \label{fig:sim_vary_d}
\end{figure}

It is interesting to examine how MCAP behaves when there is no covariance signal, and when the data are generated from a simple, K-means-like model. 
\autoref{fig:sim_vary_d} considers this case, showing adjusted Rand Index as a function of the mean distance parameter $d$ for two simulation settings with $K=2$ and $K=4$, respectively, where the groups are sampled from isotropic Gaussian distributions.

\FloatBarrier

\section*{Appendix B}
\label{app:sim_sigma}
\begin{sidewaysfigure}[ht]
	\centering    
    \includegraphics[width=1.0\textwidth]{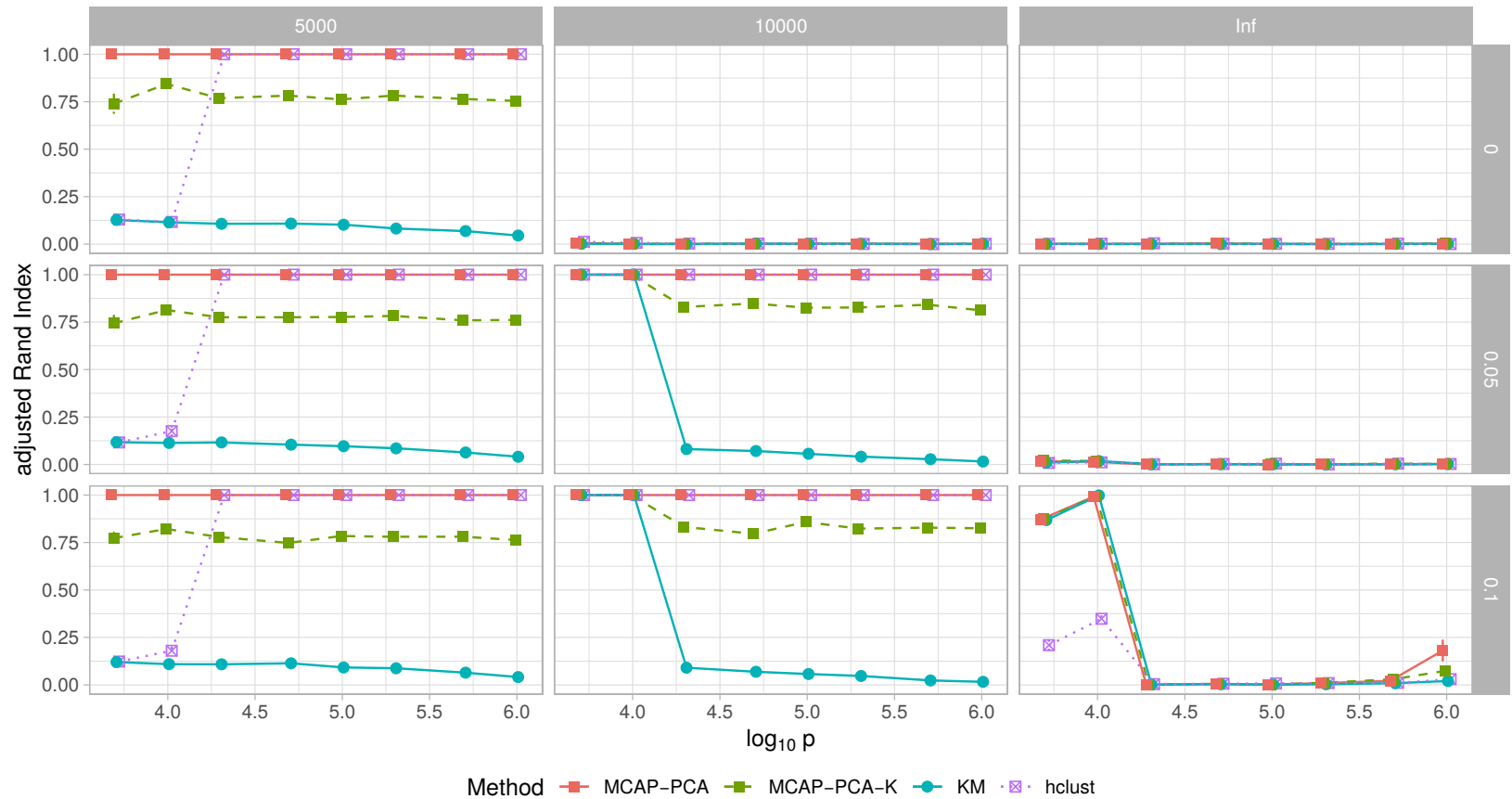}
    \caption[Simulation results: varying $p$ for different settings of Wishart $dof$ and $d$.]{Simulated data to assess performance when varying the covariance signal (using a subsampling for settings with $p>10^4$, see text for further details on how the data was generated). The y-axis shows the dimensionality $p$ on a $\log_{10}$-scale. Columns show different settings of the Wishart degrees of freedom (indicated at the top) and rows show three scenarios for different mean signal $d$ (indicated on the side); with $K=2$, $n_k = 100$. Errorbars indicate the standard error of the mean over 10 realizations (mostly smaller than actual plotting symbols.}
    \label{fig:sim_vary_sigma}
\end{sidewaysfigure}

\autoref{fig:sim_vary_sigma} shows additional experiments with regard to varying the covariance signal (see also \autoref{fig:sim_vary_sigma_full}). Due to computational complexity, simulating high dimensional covariance matrices from an inverse Wishart distribution, as well as sampling very high dimensional data from a multivariate Gaussian, is cumbersome. We therefore considered a re-sampling approach for settings where $p>10^4$. As in the original setup, we draw blocks of size $B=5{,}000$ from an inverse Wishart distribution to form the basis for the covariance matrix. We then sample a  base matrix $X^{(0)}$ of size $10^4 \times 10^4$, using two identical blocks to form a block-diagonal covariance matrix. For any $p>10^4$ we concatenate randomly drawn columns and, for each column, $n_k$ randomly subsampled entries, up to $p=10^6$. This allows for much more efficient sampling in very high dimensions. However, the subsampling means that the full covariance structure is no longer explicitly accessible precluding direct computation of the Bayes risk.

\FloatBarrier

\end{document}